# RadioRAG: Online Retrieval-augmented Generation for Radiology Question Answering

Soroosh Tayebi Arasteh (1,2,3,4), Mahshad Lotfinia (1), Keno Bressem (5,6), Robert Siepmann (1), Lisa Adams (6), Dyke Ferber (7,8), Christiane Kuhl (1), Jakob Nikolas Kather (7,8,9), Sven Nebelung (1), Daniel Truhn (1)

(1) Department of Diagnostic and Interventional Radiology, University Hospital RWTH Aachen, Aachen, Germany.
(2) Pattern Recognition Lab, Friedrich-Alexander-Universität Erlangen-Nürnberg, Erlangen, Germany.
(3) Department of Urology, Stanford University, Stanford, CA, USA.
(4) Department of Radiology, Stanford University, Stanford, CA, USA.
(5) Institute for Radiology and Nuclear Medicine, German Heart Centre Munich, Technical University of Munich, Munich, Germany.
(6) Department of Radiology, Klinikum rechts der Isar, Technical University of Munich, Munich, Germany.
(7) Else Kroener Fresenius Center for Digital Health, Faculty of Medicine and University Hospital Carl Gustav Carus, TUD Dresden University of Technology, Dresden, Germany
(8) Medical Oncology, National Center for Tumor Diseases (NCT), University Hospital Heidelberg, Heidelberg, Germany.
(9) Department of Medicine I, Faculty of Medicine and University Hospital Carl Gustav Carus, TUD Dresden University of Technology, Dresden, Germany.

**Correspondence**
Soroosh Tayebi Arasteh, Dr.-Ing., Dr. rer. medic.
Department of Diagnostic and Interventional Radiology,
University Hospital RWTH Aachen
Pauwelsstr. 30
52074 Aachen, Germany
soroosh.arasteh@rwth-aachen.de

**Abstract**

Large language models (LLMs) often generate outdated or inaccurate information based on static training datasets. Retrieval-augmented generation (RAG) mitigates this by integrating outside data sources. While previous RAG systems used pre-assembled, fixed databases with limited flexibility, we have developed Radiology RAG (RadioRAG), an end-to-end framework that retrieves data from authoritative radiologic online sources in real-time. We evaluate the diagnostic accuracy of various LLMs when answering radiology-specific questions with and without access to additional online information via RAG. Using 80 questions from the RSNA Case Collection across radiologic subspecialties and 24 additional expert-curated questions with reference standard answers, LLMs (GPT-3.5-turbo, GPT-4, Mistral-7B, Mixtral-8x7B, and Llama3 [8B and 70B]) were prompted with and without RadioRAG in a zero-shot inference scenario RadioRAG retrieved context-specific information from [www.radiopaedia.org](www.radiopaedia.org) in real-time. Accuracy was investigated. Statistical analyses were performed using bootstrapping. The results were further compared with human performance. RadioRAG improved diagnostic accuracy across most LLMs, with relative accuracy increases ranging up to 54% for different LLMs. It matched or exceeded non-RAG models and the human radiologist in question answering across radiologic subspecialties, particularly in breast imaging and emergency radiology. However, the degree of improvement varied among models; GPT-3.5-turbo and Mixtral-8x7B-instruct-v0.1 saw notable gains, while Mistral-7B-instruct-v0.2 showed no improvement, highlighting variability in RadioRAG's effectiveness. LLMs benefit when provided access to domain-specific data beyond their training data. RadioRAG shows potential to improve LLM accuracy and factuality in radiology question answering by integrating real-time domain-specific data.

# 1. Introduction

Artificial intelligence (AI) is in the process of changing diagnostic radiology by enhancing image analysis, improving diagnostic accuracy, and streamlining workflow processes[1]. Recent advances in large language models (LLMs)[2–4] have demonstrated potential in extracting structured information from radiological reports[5,6], enhancing data mining capabilities[7], improving diagnostic accuracy[6,8], and enabling more reliable speech recognition[9]. However, the use of LLMs in radiology comes with challenges, most prominently the risk of generating inaccurate information and perpetuating biases[10–12]. Strategies like human feedback[13] and prompt engineering have been employed to refine outputs but ultimately cannot solve the problem[1,14,15]. This is due to the fact that LLMs have to rely on their internal knowledge which is incomplete and may be biased. Rather, it was proposed that LLMs should be used as reasoning engines[16] with access to external sources that they can access. This approach is called retrieval-augmented generation (RAG)[17] and may remedy two problems: firstly, the risk of hallucinating information is reduced, since source material can be used and cited[18]. Secondly, LLMs can access up-to-date information through RAG, while conventional LLM querying has to rely on the information fed to the model during training.

Recent studies have demonstrated the effectiveness of RAG in answering general clinical questions[19,20]. However, its application in radiology has not been explored. In this study, we introduce Radiology RAG (RadioRAG) as a novel framework tailored specifically for typical inquiries in diagnostic radiology.

RadioRAG employs LLMs as reasoning engines to process user questions. It determines which external sources to query for relevant information, collects the source data, and then compiles a comprehensive answer for the user. Most existing solutions that employ RAG use static, pre-compiled literature databases[19,20]. In contrast, RadioRAG accesses up-to-date information from radiopaedia[21] to collect its source data. For more information about Radiopaedia's update frequency, please visit https://radiopaedia.org/terms. This architecture enables real-time gathering of contextually relevant information and constructing the database. To our knowledge, RadioRAG is the first implementation of this paradigm in radiology. The hypothesis that we investigated were: 1) the real-time context retrieval system reduces the occurrence of hallucinations and 2) RadioRAG improves the accuracy of LLM responses to detailed questions.

# 2. Materials and Methods

This retrospective study was conducted in compliance with the Declaration of Helsinki and the relevant guidelines and regulations. The study protocol was approved by the Institutional Review Board (IRB) of the Medical Faculty of RWTH Aachen University (No. EK 028/19).

## 2.1. RSNA Cases

The existing datasets for medical (QA) answering, such as MultiMedQA[11], MedMCQA[22], and PubMedQA[23], focus on general medicine and do not cater to the specific needs of diagnostic radiology. To address this gap, we created a tailored dataset, RSNA-RadioQA, using 80 peer-reviewed cases from the Radiological Society of North America (RSNA) Case Collection (https://cases.rsna.org/). STA, ML, and DT, with 6, 2, and 14 years of experience, respectively, curated the RSNA-RadioQA dataset. Our dataset covers 18 radiology subspecialties, with at least 5 cases per subspecialty in most cases, prioritizing the most recently published cases. Questions were created by providing the clinical history from the RSNA's case description along with the image characteristics as described in the figure caption. Since we concentrated on LLM without image processing capabilities, the image itself was not provided. Care was taken to exclude any differential diagnoses provided for the case. **Figure 1** illustrates a typical example for such a question. **Table 1** provides detailed information on the full RSNA-RadioQA dataset. We make this dataset available as open-source in **Appendix S1**.

## 2.2. Expert-Curated Cases

Data contamination is a significant challenge that arises when LLMs are trained on widely sourced web data that might include the datasets used for their evaluation[19,24]. Although solutions like ClinicalQA[19] have attempted to address these gaps in the general medical domain, a radiology-specific dataset had been lacking.

We therefore developed an additional dataset of 24 typical questions in radiology that we call ExtendedQA. These questions were carefully crafted by a radiologist (RS with 5 years of experience and a specialty in diagnostic and interventional radiology). The questions were validated by another board-certified radiologist (DT with 14 years of experience and a specialty in diagnostic and interventional radiology). The complete ExtendedQA dataset is available as open-access in **Appendix S2**.

## 2.3. System Design

**Figure 2** gives an overview over the design of RadioRAG in an end-to-end framework. The following sections detail each component of the process. A glossary of key technical terms can be found in **Appendix S3**.

**Table 1: Characteristics of the datasets used in the study.** For more details about the RSNA-RadioQA and ExtendedQA datasets, refer to **Appendices S1** and **S2**. *The youngest patient was a 2-day old baby. SD: Standard deviation, N/A: Not available.

| Dataset | RSNA-RadioQA | ExtendedQA |
|---|---|---|
| **Patient age [years]** | | |
| Median | 44 | N/A |
| Mean ± SD | 44 ± 21 | |
| Range | (0*, 80) | |
| **Patient sex [n (%)]** | | |
| Total | 80 (100%) | N/A |
| Female | 37 (46%) | |
| Male | 43 (54%) | |
| **Number of questions per subspecialty [n (%)]** | | |
| Total | 80 (100%) | 24 (100%) |
| Breast Imaging | 10 (12%) | 0 (0%) |
| Cardiac | 8 (10%) | 2 (8%) |
| Chest | 13 (16%) | 7 (29%) |
| Computed Tomography | 28 (35%) | 7 (29%) |
| Emergency Radiology | 6 (8%) | 3 (13%) |
| Gastrointestinal | 12 (15%) | 6 (25%) |
| Genitourinary | 8 (10%) | 1 (4%) |
| Head and Neck | 9 (11%) | 1 (4%) |
| Magnetic Resonance Imaging | 20 (25%) | 7 (29%) |
| Molecular Imaging | 11 (14%) | 0 (0%) |
| Musculoskeletal | 14 (18%) | 6 (25%) |
| Neuroradiology | 11 (14%) | 0 (0%) |
| Nuclear Medicine | 13 (16%) | 0 (0%) |
| Oncologic Imaging | 16 (20%) | 0 (0%) |
| Pediatric | 7 (9%) | 1 (4%) |
| Radiation Oncology | 9 (11%) | 0 (0%) |
| Ultrasound | 10 (12%) | 0 (0%) |
| Vascular Imaging | 13 (16%) | 3 (13%) |

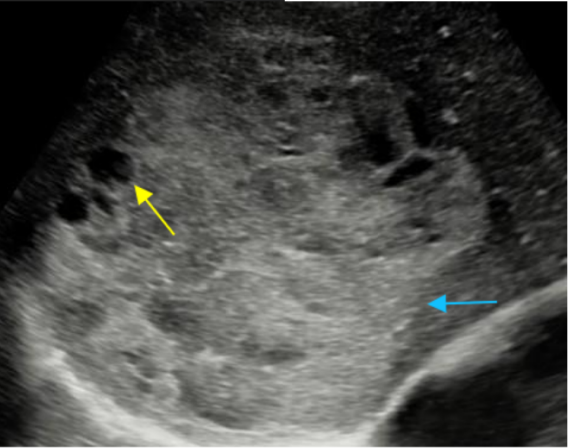

**Image/Video details**

**Figure legend:**
Axial proton-density weighted MR image with fat suppression through the right thigh above the level of the knee shows a hyperintense 13 cm mass with multiple low-intensity internal septations (blue arrow) deep to the vastus musculature (yellow arrow). The mass is anterior to and partially encasing the distal femoral diaphysis (white arrow). There is no cortical disruption to suggest osseous invasion.

**- Clinical information**

**Age and gender:** 56 year old male
**Clinical history & presentation:** 6-year-old male presented with 3-months history of off and on fever relieved by medication, weight loss, and constipation.
**Pathology:** Ultrasound guided biopsy was performed using 18G and 16 cm needle and sample sent for the histopathological analysis. Macroscopic appearance revealed well circumscribed mass showing cystic -necrotic areas containing hemorrhage as well as myxoid matrix. Microscopic examination revealed atypical spindle like cells showing repetitive mitotic figures and positivity for vimentin staining.
**Physical exam:** On physical examination, the liver was palpable 5-6 cm below the costal margin, firm in consistency, and moving proportionately with respiration. No redness over right hepatic region. Localized tenderness present. The spleen was not palpable. There were no palpable lymph nodes.
**Other diagnostic testing:** Serum AFP levels were <1.3 ng/ml.

▼

**LLM**  RSNA-RadioQA-Q37

You are a helpful expert medical research assistant. Answer the following question. Use one sentence only and keep the answer concise:

**Question:** A 6-year-old male presented with a 3-month history of off-and-on fever relieved by medication, weight loss, and constipation. Ultrasound-guided biopsy was performed using 18G and 16 cm needle and the sample was sent for histopathological analysis. Macroscopic appearance revealed a well-circumscribed mass showing cystic-necrotic areas containing hemorrhage as well as a myxoid matrix. Microscopic examination revealed atypical spindle-like cells showing repetitive mitotic figures and positivity for vimentin staining. A transverse grey-scale ultrasound image of the abdomen in a supine position shows a large heterogeneously hyperechoic mass. It occupies the entire right lobe of the liver involving segments V, VI, VII, and VIII. The mass shows hyperechoic and anechoic cystic components within. What is the most likely diagnosis?

**Figure 1: RSNA-RadioQA dataset generation.** The image is shown for context only; no images were included in this study. The screenshot is taken from a peer-reviewed case from RSNA Case Collection in[25], with the question ID: RSNA-RadioQA-Q37.

### *2.3.1. Browser and the Database*

To investigate the reasoning capabilities of the LLMs in our study, we isolated automated query generation and matching to external sources by using the same model for all LLMs: using GPT-3.5-turbo via its API, the system extracts up to five search key-phrases from a given radiological question. The threshold of five was chosen experimentally, as we observed that relevant sources were always found within the first five searches of radiopaedia, minimizing unnecessary searches and reducing system response time. The prompt specified to GPT-3.5-turbo is: "You are a helpful expert medical research assistant. I have a medical question, particularly in the field of radiology. Please summarize the question to extract the most representative keywords for use in online scientific article searches. Return a maximum of five keywords that are scientifically relevant to radiology." Two examples were provided to the model within the prompt, as shown in **Figure 2B** (two-shot approach). For each question, the same set of key-phrases was used for all RadioRAG-powered LLMs to ensure a fair comparison.

After acquiring relevant search keywords, the model searches through articles on www.radiopaedia.org, selecting the five articles most pertinent to each keyword. As part of our validation process, STA and DT reviewed the final selected articles to ensure their appropriateness. These articles are then segmented into chunks of 1,000 tokens, each with a 200 token overlap. Each chunk is converted into a vector using the 'text-embedding-ada-002' embedding function from OpenAI and temporarily stored in a vector database managed by Chroma (https://www.trychroma.com/).

### 2.3.2. Retriever

With the database prepared, the original query is also embedded into a vector using the same embedding function. This query vector is then compared against all vectors in the database using cosine similarity to retrieve the top three most similar vectors (k=3). These vectors are matched to their textual form, and the relevant text is prepared for the next stage. The LangChain framework is used for this retrieval process.

### 2.3.4. Large Language Model (LLM)

The final stage involves the respective LLM under investigation, which receives the original query along with the contextually relevant text fragments retrieved in the previous step. The LLM is instructed to provide a concise answer in one sentence, based solely on the provided context. If the answer is unknown, the LLM must explicitly state this. We used the following prompt: "Use the following pieces of retrieved context to answer the question. If you don't know the answer, say 'I don't know.' Answer concisely in one sentence." This process contrasts with traditional LLM QA methods that involve responding to queries without additional context, typically prompted with: "You are a helpful expert medical research assistant. Answer the following question concisely in one sentence". To ensure reproducibility, a temperature value of 0 was set for all LLM responses, except for those involving Mistral and Mixtral models, where a minimum temperature of 0.1 was necessary (for which we set it at 0.1). A top-p value of 1 was consistently used across all cases. Through this choice, provided results were reproducible for the same model version.

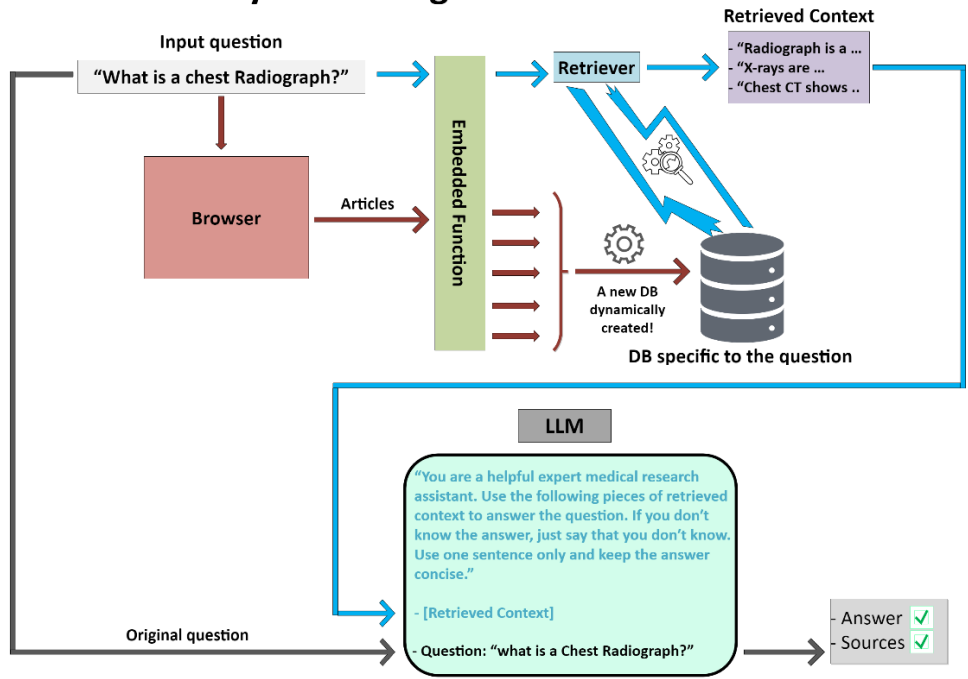

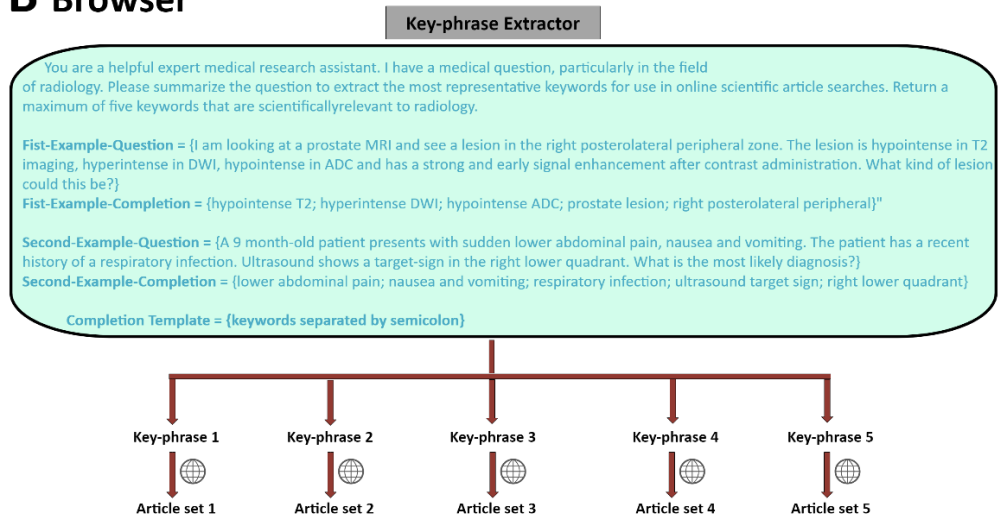

**Figure 2: Radiology Retrieval Augmented Generation (RadioRAG) architecture overview. (A)** Shows that RadioRAG system design. Initially, the input question is analyzed and the relevant articles are retrieved using the "Browser" module, which are then chunked into multiple documents. These documents are converted into vector representations and stored in a dynamically created, on-the-fly vector database (DB) for each query. The question is embedded using the same function and its vector compared against the DB vectors using cosine distance to select the closest three vectors. These vectors are reverted to their textual form to create the context for the final step. The LLM then receives the original question along with this context and is directed to use the context to formulate an answer. **(B)** Shows the architecture of the Browser module. After analyzing the input question, the 5 most representative key-phrases are extracted using the key-phrase extractor, implemented with GPT-3.5-turbo in our examples. Each key-phrase triggers the retrieval of relevant articles from www.radiopaedia.org.

## 2.4. Evaluation

STA and DT performed the evaluation. To assess the efficacy of RadioRAG across varying scales of language models, we tested both smaller and larger LLMs. All final LLM responses for this study were generated between April 1 and April 25, 2024. We included GPT-3.5-turbo, GPT-4, Mistral-7B-instruct-v0.2, Mixtral-8x7B-instruct-v0.1, Llama3-8B, and Llama3-70B-instruct. This set of models represents the state-of-the-art in size and capabilities. Each model was integrated into the RadioRAG pipeline and evaluated in both a conventional QA setup and within the RadioRAG framework.

The performance of all models was evaluated by comparing their responses within the RadioRAG framework to those in conventional QA, using reference standard answers as the benchmark. Although various metrics like naturalness, fluency, and coherence are commonly used in LLM evaluation[26–28], we prioritized accuracy[19,29,30] as the primary metric due to the specific nature of our application, which demands correct and concise diagnostic answers. Accuracy was measured by scoring responses as true (1) if they correctly addressed the query and false (0) otherwise[19]. Factuality was also assessed by verifying the suitability of the sources the LLMs cited for each answer. Additionally, the models' ability to recognize and admit when the available information was insufficient was evaluated, requiring them to state "I don't know" in such instances. Any inaccuracies or omissions in expressing uncertainty were considered deviations from expected factuality and transparency.

Moreover, to assess the practical utility of RadioRAG, we compared the diagnostic performance of RadioRAG-powered LLMs to human radiologist on both the RSNA-RadioQA and ExtendedQA datasets. A board-certified radiologists (LA with 9 years of experience and subspecialty in chest and oncologic imaging) was blinded to the reference standard answers, curation of the datasets, and LLM responses. The human expert answered the same set of questions given to the LLMs, based solely on their own knowledge and without accessing any online information or additional materials such as images. The responses were evaluated by STA and DT using the same accuracy metrics applied to the LLMs.

## 2.5. Statistical Analysis

Statistical analysis was conducted using Python v3.11 with SciPy v1.11, NumPy v1.24, and statsmodels v0.14. packages. To evaluate the variability, separately for each dataset, bootstrapping was employed with 10,000 redraws for the metrics to determine the mean, standard deviation, 95% confidence intervals (CI), and to calculate p-values for differences in accuracy between the RadioRAG and non-RadioRAG setups[31], ensuring a strictly paired setup where redraws were identical across conditions[32]. To account for multiple comparisons, the p-values were adjusted for multiplicity using the false discovery rate (FDR). An FDR<0.05 was used.

## 2.6. Code and Data Availability

All source codes and datasets used in this study are publicly available to ensure transparency and reproducibility. The code has been developed using Python v3.11 with PyTorch v2.1 and is hosted on GitHub at https://github.com/tayebiarasteh/RadioRAG. We utilized the LangChain v0.1.0 framework for the RadioRAG pipeline, with Chroma serving as the vector database. The OpenAI API v1.12 provided access to the GPT-4 and GPT-3.5-turbo models as well as the 'text-embedding-ada-002' embedding function from OpenAI. Additionally, the Replicate API v0.25 facilitated cloud execution of Mistral and Mixtral models without local GPU requirements, and Ollama (https://ollama.com/) framework for execution of the newly-released Llama3 open-source models. The underlying datasets are publicly accessible and included in the supplemental materials of this publication.

# 3. Results

## 3.1. Dataset Characteristics

Mean age over all patients in the RSNA-RadioQA dataset was 44 ± [SD] 21 years, with a range of 2 days to 80 years. **Table 1** reports the characteristics of the dataset and the distribution of subspecialties among all questions.

## 3.2. Impact of RadioRAG on Diagnostic Performance of LLMs

Typical responses of the LLMs to an exemplary question from the RSNA-RadioQA dataset are given in **Table 2.** RadioRAG increased the accuracy of the LLMs' responses on the ExtendedQA dataset as illustrated by **Figure 3**. Detailed results are given in **Table 3**: the accuracy of GPT-3.5-turbo increased from 66% ± 5 (53/80) to 74% ± 5 (59/80) (P=0.03), of GPT-4 from 78% ± 5 (62/80) to 79% ± 5 (63/80) (P=0.28), of Mixtral-8x7B-instruct-v0.1 from 65% ± 5 (52/80) to 76% ± 5 (61/80) (P=0.02), of Llama3-8B from 58% ± 6 (46/80) to 59% ± 5 (47/80) (P=0.39), and of Llama3-70B from 66% ± 5 (53/80) to 69% ± 5 (55/80) (P=0.30). The only exception was Mistral-7B-instruct-v0.2, which exhibited no change (55% ± 6 (44/80) in both cases; P=0.45). Similarly, for the ExtendedQA dataset, all LLMs demonstrated improvements. In particular, Mixtral-8x7B-instruct-v0.1, Llama3-8B, and Mistral-7B all exhibited significant improvements, while the bigger models also showed improvements, yet without reaching the significance threshold. Detailed results stratified along subspecialties are given in **Table S1.**

## 3.3. Open-Weights Models Benefit from RadioRAG

While RadioRAG consistently improved diagnostic performance across all tested LLMs, the degree of improvement varied significantly between models. The most complex model, GPT-4 exhibited relative accuracy improvements of 1% [(79-78)/78] and 6% [(75-71)/71], respectively for the two QA datasets, while GPT-3.5-turbo exhibited relative accuracy improvements of 12% [(74-66)/66] and 22% [(71-58)/58], respectively. We observed stronger increases on the ExtendedQA dataset for the open-weights models Mistral-7B-instruct-v0.2 (up to 54% [(71-46)/46], Mixtral-8x7B-instruct-v0.1 (up to 47% [(79-54)/54] and Llama3-8B (up to 34% [(67-50)/50]. Importantly, while open-weights models had inferior performance in the non-RAG setting, RadioRAG rendered these models competitive with GPT-4.

## 3.4. RadioRAG Enforces Factuality in LLMs

We found that RadioRAG guided the LLMs to ground their answers in factual content from the source data, i.e., whether the provided answer is based on and related to the retrieved context[19]. **Table S2** presents detailed quantitative results on hallucination rates for each model. **Tables S3** and **S4** provide overviews of questions, with and without RAG, respectively, using GPT-3.5-turbo as an example.

Following the review of the retrieved articles, we generally found that relevant articles were selected in 72% (58/80) of questions for the RSNA-RadioQA dataset and 83% (20/24) for the ExtendedQA dataset. However, in some cases—possibly when a related article was not available on Radiopaedia—unrelated articles were chosen. Additionally, while answers typically aligned closely with the source data, strict adherence occasionally led to inaccuracies when the retrieved articles were not fully relevant to the query, affecting between 15% (12/80) and 20% (16/80) of questions in the RSNA-RadioQA dataset across different LLMs, and between 8% (2/24) and 12% (3/24) of questions in the ExtendedQA dataset for all LLMs. **Table 4** provides an example where the enforcement of factuality resulted in an incorrect answer due to irrelevant context.

This enforcement generally minimized hallucinations by the LLMs; however, different LLMs exhibited varying behavior. For both datasets, Mixtral-8x7B-instruct-v0.1 and GPT-4 showed the fewest hallucinations, with 9% (7/80) and 6% (5/80) for the RSNA-RadioQA dataset, respectively, and 8% (2/24) and 12% (3/24) for the ExtendedQA dataset, respectively.

## 3.5. Comparison to Human Performance

The human expert radiologist achieved an accuracy of 63% ± 5 (50/80) (95% CI: 51%, 72%) on the RSNA-RadioQA dataset, which was significantly lower than RadioRAG-powered models of GPT-3.5-turbo (P=0.007), GPT-4 (P=0.001), and Mixtral-8x7B-instruct-v0.1 (P=0.007), but not significantly different from Llama3-70B (P=0.21). It outperformed RadioRAG-powered models of Mistral-7B-instruct-v0.2 (P=0.22) and Llama3-8B (P=0.31). On the ExtendedQA dataset, the radiologist achieved 62% ± 10 (15/24) (95% CI: 42%, 83%), which was lower than all RadioRAG-powered models, though the differences were not statistically significant (P>0.19).

## A RSNA-RadioQA Dataset

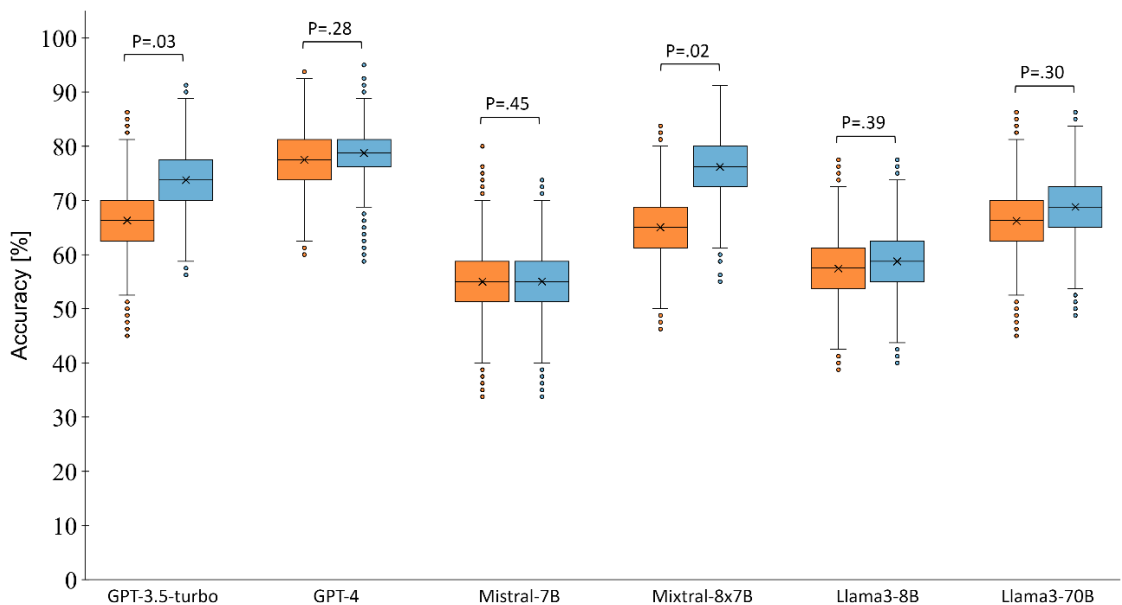

## B ExtendedQA Dataset

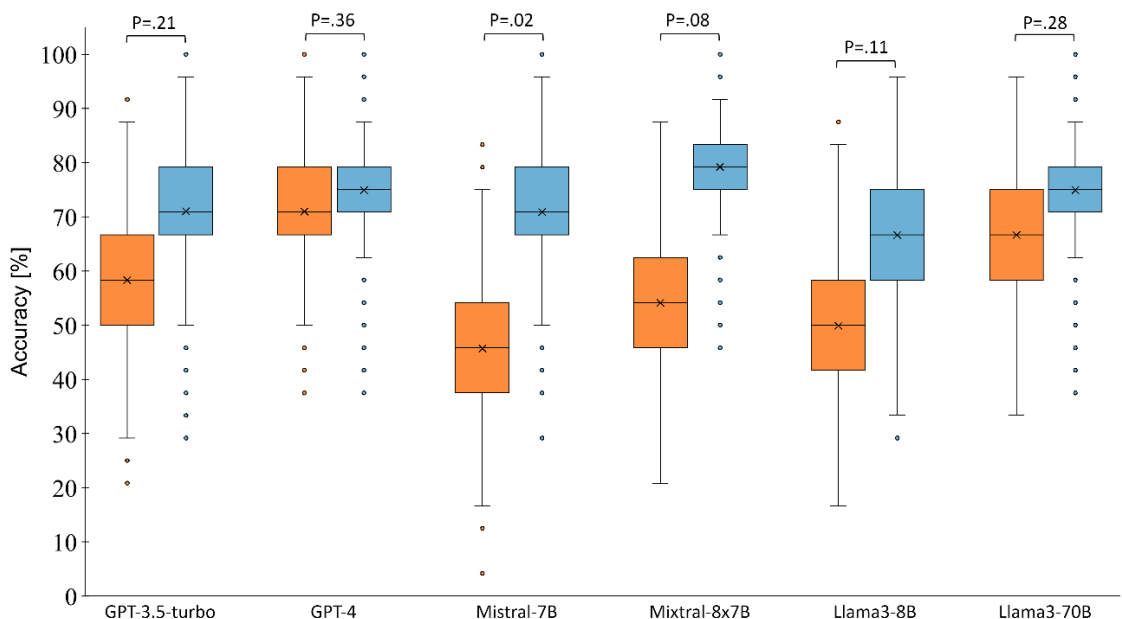

**Figure 3: Quantitative evaluation of RadioRAG across datasets.** This figure displays the accuracy results on two datasets: **A)** RSNA-RadioQA with n=80 (details in **Appendix S1**) and **B)** ExtendedQA with n=24 (details in **Appendix S2**). The LLMs included in the evaluation are GPT-3.5-turbo, GPT-4, Mistral-7B-instruct-v0.2 (Mistral-7B), Mixtral-8x7B-instruct-v0.2 (Mixtral-8x7B), Llama3-8B, and Llama3-70B-instruct (Llama3-70B). The orange boxes correspond to the models without using RadioRAG, while the blue boxes correspond to the RadioRAG-powered models. The analysis employs bootstrapping with 10,000 repetitions, allowing replacements. P-values were calculated between each of the RadioRAG-powered methods and their non-RadioRAG counterpart. A value below 0.05 was considered significant.

# 4. Discussion

In this study, we introduced Radiology RAG (RadioRAG), a novel framework that enhances the diagnostic accuracy of LLMs by utilizing contextually relevant data from an established radiological source. To benchmark RadioRAG, we developed two datasets which we make publicly available: RSNA-RadioQA for internal testing and ExtendedQA as an external dataset.

Overall, our findings show that RadioRAG-powered LLMs outperformed their non-RAG counterparts in most cases, yielding more accurate and reliable outputs in radiological contexts.

To evaluate RadioRAG, we developed RSNA-RadioQA, a diagnostic radiology QA dataset using the RSNA Case Collection. This collection includes peer-reviewed cases from various global institutions and provides a diverse and representative dataset of radiological question answering across multiple subspecialties. However, as these cases had already been published online, there is a potential bias since the LLMs might have previously accessed parts of this data during their training. To mitigate this, we created the ExtendedQA dataset with 24 previously unseen questions as an additional test dataset to confirm the results on RSNA-RadioQA.

In our study, LLMs powered by RadioRAG generally outperformed those in conventional QA setups. GPT-4 excelled in conventional QA settings, but exhibited only slight improvements with RadioRAG. In contrast, LLMs like Mixtral-8x7B-instruct-v0.1, which initially lagged behind in the conventional QA setting, saw substantial gains with RadioRAG, matching or even surpassing GPT-4. This improvement is significant for two key reasons: firstly, it was achieved without additional training, relying solely on strategic guidance during inference—a cost-effective method compared to full model retraining. Secondly, it suggests that high-performance natural language processing may be achievable using open-source LLMs with RAG, potentially reducing reliance on closed-source models like GPT-4.

Previous studies[19,20,33,34] have applied RAG to medical question answering, yet none have specifically focused on radiology, and except for[19], most have utilized offline RAG where documents from books, articles, and websites are compiled into a static database for retrieval. This approach lacks the immediacy of an online RAG like RadioRAG. By accessing up-to-date information, RadioRAG can adapt to changes in guidelines, image assessment strategies, or updates to imaging techniques.

**Table 2: LLM responses to one of the questions, with and without RadioRAG.** This exemplary question is chosen from the RSNA-RadioQA dataset (details in **Appendix S1**) with the question ID: RSNA-RadioQA-Q37.

| | |
|---|---|
| Question | A 6-year-old male presented with a 3-month history of off-and-on fever relieved by medication, weight loss, and constipation. Ultrasound-guided biopsy was performed using 18G and 16 cm needle and the sample was sent for histopathological analysis. Macroscopic appearance revealed a well-circumscribed mass showing cystic-necrotic areas containing hemorrhage as well as a myxoid matrix. Microscopic examination revealed atypical spindle-like cells showing repetitive mitotic figures and positivity for vimentin staining. A transverse grey-scale ultrasound image of the abdomen in a supine position shows a large heterogeneously hyperechoic mass. It occupies the entire right lobe of the liver involving segments V, VI, VII, and VIII. The mass shows hyperechoic and anechoic cystic components within. What is the most likely diagnosis? |
| Reference Standard Answer | Undifferentiated embryonal sarcoma |
| RadioRAG Pipeline Key-phrases Source | **Key-phrases:** fever, weight loss, constipation, ultrasound-guided biopsy, liver mass<br>**Most relevant source among the 25 retrieved articles:**<br>https://radiopaedia.org/articles/undifferentiated-embryonal-sarcoma-of-the-liver?lang=us |
| Conventional GPT-3.5-turbo | The most likely diagnosis is hepatoblastoma |
| RadioRAG+GPT-3.5-turbo | The most likely diagnosis is undifferentiated embryonal sarcoma of the liver. |
| Conventional GPT-4 | The most likely diagnosis for the 6-year-old male with these symptoms and ultrasound findings is a hepatic sarcoma, specifically undifferentiated embryonal sarcoma of the liver (UESL). |
| RadioRAG+GPT-4 | The most likely diagnosis is an undifferentiated embryonal sarcoma of the liver. |
| Conventional Mistral-7B | The most likely diagnosis is a malignant peripheral nerve sheath tumor (MPNST) or neurofibrosarcoma. |
| RadioRAG+Mistral-7B-instruct-v0.2 | The most likely diagnosis is an undifferentiated embryonal sarcoma of the liver. |
| Conventional Mixtral-8x7B-instruct-v0.1 | The most likely diagnosis is hepatoblastoma, a malignant pediatric liver tumor characterized by atypical spindle-like cells, cystic-necrotic areas, and vimentin positivity, as described in the patient's presentation and ultrasound findings. |
| RadioRAG+Mixtral-8x7B-instruct-v0.1 | The most likely diagnosis is an undifferentiated embryonal sarcoma of the liver, given the patient's age, symptoms, and the ultrasound and histopathological findings. |
| Conventional Llama3-8B | The most likely diagnosis is a pediatric hepatoblastoma, a rare malignant tumor that originates in the liver, based on the histopathological features and imaging findings. |
| RadioRAG+Llama3-8B | Undifferentiated embryonal sarcoma of the liver. |
| Conventional Llama3-70B-instruct | The most likely diagnosis is Undifferentiated Embryonal Sarcoma (UES) of the liver, a rare and aggressive pediatric tumor. |
| RadioRAG+Llama3-70B-instruct | The most likely diagnosis is undifferentiated embryonal sarcoma of the liver. |

**Table 3: Statistical analysis results for the RSNA-RadioQA and ExtendedQA datasets.** P-values were calculated between each of the RadioRAG-based methods and all the other models. N/A: Not assigned.

| Model Name | Accuracy (mean ± SD [95% CI]) [%] | P-Value | | | | | |
|---|---|---|---|---|---|---|---|
| | | RAG+GPT-3.5-turbo | RAG+GPT-4 | RAG+Mistral-7B | RAG+Mixtral-8x7B | RAG+Llama3-8B | RAG+Llama3-70B |
| **RSNA-RadioQA Dataset** | | | | | | | |
| RAG+GPT-3.5-turbo | 74 ± 5 [95% CI: 64, 84] (59/80) | N/A | 0.07 | 0.001 | 0.26 | 0.005 | 0.20 |
| RAG+GPT-4 | 79 ± 5 [95% CI: 70, 88] (63/80) | | N/A | 0.001 | 0.35 | 0.001 | 0.02 |
| RAG+Mistral-7B | 55 ± 6 [95% CI: 44, 66] (44/80) | | | N/A | 0.001 | 0.24 | 0.005 |
| RAG+Mixtral-8x7B | 76 ± 5 [95% CI: 66, 85] (61/80) | | | | N/A | 0.003 | 0.09 |
| RAG+Llama3-8B | 59 ± 5 [95% CI: 47, 70] (47/80) | | | | | N/A | 0.05 |
| RAG+Llama3-70B | 69 ± 5 [95% CI: 59, 79] (55/80) | | | | | | N/A |
| GPT-3.5-turbo | 66 ± 5 [95% CI: 56, 76] (53/80) | 0.03 | 0.001 | 0.06 | 0.04 | 0.15 | 0.28 |
| GPT-4 | 78 ± 5 [95% CI: 69, 86] (62/80) | 0.26 | 0.28 | 0.001 | 0.45 | 0.001 | 0.03 |
| Mistral-7B | 55 ± 6 [95% CI: 44, 66] (44/80) | 0.001 | 0.001 | 0.45 | 0.001 | 0.26 | 0.02 |
| Mixtral-8x7B | 65 ± 5 [95% CI: 55, 75] (52/80) | 0.05 | 0.003 | 0.06 | 0.02 | 0.20 | 0.26 |
| Llama3-8B | 58 ± 6 [95% CI: 46, 69] (46/80) | 0.001 | 0.001 | 0.39 | 0.001 | 0.39 | 0.03 |
| Llama3-70B | 66 ± 5 [95% CI: 56, 76] (53/80) | 0.07 | 0.01 | 0.04 | 0.04 | 0.17 | 0.30 |
| **ExtendedQA Dataset** | | | | | | | |
| RAG+GPT-3.5-turbo | 71 ± 9 [95% CI: 50, 88] (17/24) | N/A | 0.34 | 0.44 | 0.18 | 0.44 | 0.36 |
| RAG+GPT-4 | 75 ± 9 [95% CI: 58, 92] (18/24) | | N/A | 0.44 | 0.34 | 0.21 | 0.42 |
| RAG+Mistral-7B | 71 ± 9 [95% CI: 54, 88] (17/24) | | | N/A | 0.28 | 0.44 | 0.28 |
| RAG+Mixtral-8x7B | 79 ± 8 [95% CI: 62, 96] (19/24) | | | | N/A | 0.27 | 0.44 |
| RAG+Llama3-8B | 67 ± 10 [95% CI: 46, 83] (16/24) | | | | | N/A | 0.001 |
| RAG+Llama3-70B | 75 ± 9 [95% CI: 58, 92] (18/24) | | | | | | N/A |
| GPT-3.5-turbo | 58 ± 10 [95% CI: 38, 75] (14/24) | 0.21 | 0.11 | 0.11 | 0.11 | 0.28 | 0.08 |
| GPT-4 | 71 ± 9 [95% CI: 50, 88] (17/24) | 0.44 | 0.36 | 0.44 | 0.30 | 0.44 | 0.34 |
| Mistral-7B | 46 ± 10 [95% CI: 25, 67] (11/24) | 0.10 | 0.01 | 0.02 | 0.02 | 0.08 | 0.001 |
| Mixtral-8x7B | 54 ± 10 [95% CI: 33, 75] (13/24) | 0.18 | 0.13 | 0.14 | 0.08 | 0.27 | 0.11 |
| Llama3-8B | 50 ± 10 [95% CI: 29, 71] (12/24) | 0.10 | 0.02 | 0.04 | 0.02 | 0.11 | 0.001 |
| Llama3-70B | 67 ± 10 [95% CI: 46, 83] (16/24) | 0.40 | 0.30 | 0.36 | 0.21 | 0.44 | 0.28 |

Like other RAG systems[17,19,35], RadioRAG enforces factual responses from LLMs. Despite this, we found instances where responses were incorrect. This was due to irrelevant context extracted from the sources. Such strict reliance on irrelevant source materials can lead to incorrect answers unless the LLM correctly identifies the irrelevance and states that it does not know the answer. GPT-4 and Llama3-70B-instruct were particularly adept at recognizing when they could not provide informed answers, although they still occasionally failed to do this effectively. Future research should focus on enhancing embedding functions and methodologies to retrieve more relevant context such as fine-tuned embeddings and advanced reranking[36], thus minimizing the risk of inaccuracies. Additionally, exploring agentic workflows[37], which enable dynamic adjustments based on user input and real-time feedback, could further optimize the system's performance[38], allowing the model to balance between retrieved information and its internal knowledge for improved accuracy. Moreover, future work could expand the RadioRAG framework to a multimodal paradigm where images could be included alongside textual data, and a comparison with LLMs that have dynamic web search capabilities (e.g., Perplexity AI, CA, USA) would provide insights into the advantages of controlled retrieval versus real-time web access.

RadioRAG has limitations. First, the on-the-fly generation of a database can be time-consuming, potentially extending the time it takes for RadioRAG to respond compared to conventional QA setups. A comparative analysis of the time required for QA with RadioRAG versus conventional methods is provided in **Appendix S4**. To mitigate this, we have optimized the framework to select up to 5 key-phrases and retrieve 5 articles per key-phrase, resulting in a total of 25 articles. Second, the reliance on continually querying online scientific sources, in our case www.radiopaedia.org, could overload the website, especially if multiple users are accessing it simultaneously, potentially leading to downtimes. Therefore, while RadioRAG offers a compelling proof-of-concept, more research into its efficiency and computational demands is essential before it can be implemented in clinical practice. Future deployments should consider establishing agreements with source websites to ensure fair use and manage the load effectively. Third, the small external ExtendedQA dataset (n=24) limits generalizability. While we used bootstrapping to mitigate this, larger external validations are needed, and we plan to expand the dataset in future work. Future work should validate our findings on larger external datasets. Additionally, we plan to expand the ExtendedQA dataset in future versions to enhance the robustness and generalizability of the results. Fourth, while RadioRAG is adaptable to multiple information sources, in this study, we relied exclusively on Radiopaedia, a well-established source in the radiology community. This reliance on a single source presents a potential limitation, and future studies should consider incorporating additional sources to enhance the system's versatility and accuracy.

In conclusion, RadioRAG introduces real-time data retrieval to enhance the accuracy and factuality of LLMs in radiological diagnostics. This development offers a foundation for further work that could improve diagnostic processes and patient care in healthcare.

**Table 4: Factuality and potential hallucinations in RadioRAG.** RadioRAG enforces factuality by requiring LLMs to base their responses on the content of retrieved articles. However, strict adherence to these sources can lead to inaccuracies when the context provided is not relevant to the query. In this given example, only GPT-4, powered by RadioRAG, acknowledges the inability to find a relevant answer, while the other RadioRAG-powered LLMs provide incorrect answers, demonstrating potential hallucinations. This exemplary question is chosen from the RSNA-RadioQA dataset with the question ID: RSNA-RadioQA-Q59.

| | **RadioRAG-Powered QA** |
|---|---|
| Prompt | You are a helpful expert medical research assistant. Use the following pieces of retrieved context to answer the question. If you don't know the answer, just say that you don't know. Use one sentence only and keep the answer concise: <br> *- Retrieved Context: Pathology Etiology The majority (80%) of scolioses have no apparent underlying cause and are termed idiopathic 1. Idiopathic scoliosis can be broken down into infantile, juvenile and adolescent based on the age of the patient. The remaining 20% of scolioses are the result of other causes. There are many ways to potentially group these causes, but a simple three-pronged grouping strategy is: neuromuscular: conditions that cause neurological or muscular deficits that result in asymmetric muscular tone resulting in spinal curvature congenital bony: an underlying bony abnormality of the vertebra that results in a relatively fixed spinal curve tumor or treatment: this is a bit of a catch-all for the remainder of causes, most of which relate to an adjacent tumor, or previous treatment, e.g. radiotherapy or cardiac surgery.* <br> *Causes neuromuscular scoliosis cerebral palsy Chiari malformation Friedreich ataxia syringomyelia diastematomyelia tethered cord spinal dysraphism muscular dystrophies spinal muscular atrophy connective tissue disorders (e.g. Marfan disease) traumatic paralysis congenital scoliosis segmentation and fusion abnormalities (e.g. hemivertebrae) skeletal dysplasias (e.g. achondroplasia) lumbosacral transitional vertebra tumors bone: osteoid osteoma, osteoblastoma, metastases, neurofibromas in neurofibromatosis type I soft tissues: meningioma, neurofibroma, astrocytoma, ependymoma infection resulting in bony abnormalities (e.g. pyogenic osteomyelitis, tuberculous spondylitis).* <br> *[Source: https://radiopaedia.org/articles/scoliosis]* <br> **- Question:** A 15-year-old male with no significant past medical history presents to orthopedic clinic for evaluation of scoliosis. An abnormality is seen on scoliosis films. Review of systems is positive for frequent nosebleeds. Frontal and lateral views of the spine show dextroconvex thoracic scoliosis with an incidental nodule in the right middle lobe. What is the most likely diagnosis? |
| Reference Answer | Pulmonary varix |
| RadioRAG+ GPT-4 | **Answer:** The text does not provide enough information to determine the most likely diagnosis for the 15-year-old male patient. |
| RadioRAG+ Mixtral-8x7B | **Answer:** Based on the information provided, the most likely diagnosis is idiopathic scoliosis with an incidental finding of a nodule in the right middle lobe, possibly indicative of a pulmonary condition unrelated to the scoliosis. The frequent nosebleeds may be a clue to this separate condition. However, a definitive diagnosis would require further investigation by a healthcare professional. |
| RadioRAG+ Llama3-70B | **Answer:** The most likely diagnosis is neurofibromatosis type 1 (NF1) due to the presence of a spinal curvature and an incidental pulmonary nodule, which are both common features of this genetic disorder. |
| | **Conventional QA** |
| Prompt | You are a helpful expert medical research assistant. Answer the following question. Use one sentence only and keep the answer concise.: <br> **- Question:** A 15-year-old male with no significant past medical history presents to orthopedic clinic for evaluation of scoliosis. An abnormality is seen on scoliosis films. Review of systems is positive for frequent nosebleeds. Frontal and lateral views of the spine show dextroconvex thoracic scoliosis with an incidental nodule in the right middle lobe. What is the most likely diagnosis? |
| Conventional GPT-4 | **Answer:** The most likely diagnosis for a 15-year-old male with dextroconvex thoracic scoliosis and an incidental nodule in the right middle lobe, who also experiences frequent nosebleeds, is hereditary hemorrhagic telangiectasia (HHT), also known as Osler-Weber-Rendu syndrome, which is associated with telangiectasias and arteriovenous malformations (AVMs) that can cause pulmonary nodules and epistaxis. |
| Conventional Mixtral-8x7B | **Answer:** 15-year-old male with scoliosis, nosebleeds, and a lung nodule may have neurofibromatosis type 1, a genetic disorder often associated with these symptoms. |
| Conventional Llama3-70B | **Answer:** The most likely diagnosis is Osler-Weber-Rendu syndrome, a rare genetic disorder characterized by abnormal blood vessel formation and frequent nosebleeds, which can be associated with pulmonary arteriovenous malformations and scoliosis. |

*Funding Sources*

STA was funded and partially supported by the Radiological Cooperative Network (RACOON) under the German Federal Ministry of Education and Research (BMBF) grant number 01KX2021. JNK is supported by the German Cancer Aid (DECADE, 70115166), the German Federal Ministry of Education and Research (PEARL, 01KD2104C; CAMINO, 01EO2101; SWAG, 01KD2215A; TRANSFORM LIVER, 031L0312A; TANGERINE, 01KT2302 through ERA-NET Transcan; Come2Data, 16DKZ2044A; DEEP-HCC, 031L0315A), the German Academic Exchange Service (SECAI, 57616814), the German Federal Joint Committee (TransplantKI, 01VSF21048) the European Union's Horizon Europe and innovation programme (ODELIA, 101057091; GENIAL, 101096312), the European Research Council (ERC; NADIR, 101114631), the National Institutes of Health (EPICO, R01 CA263318) and the National Institute for Health and Care Research (NIHR, NIHR203331) Leeds Biomedical Research Centre. The views expressed are those of the author(s) and not necessarily those of the NHS, the NIHR or the Department of Health and Social Care. This work was funded by the European Union. Views and opinions expressed are however those of the author(s) only and do not necessarily reflect those of the European Union. Neither the European Union nor the granting authority can be held responsible for them. SN and DT were supported by grants from the Deutsche Forschungsgemeinschaft (DFG) (NE 2136/3-1, LI3893/6-1, TR 1700/7-1). DT is supported by the German Federal Ministry of Education (TRANSFORM LIVER, 031L0312A; SWAG, 01KD2215B) and the European Union's Horizon Europe and innovation programme (ODELIA [Open Consortium for Decentralized Medical Artificial Intelligence], 101057091).

*Author Contributions*

STA and DT designed the study and performed the formal analysis. STA and DT analyzed and controlled the data. The manuscript was written by STA, ML, and DT. The experiments were performed by STA. The software was developed by STA. The statistical analyses were performed by STA and DT. The RSNA-RadioQA dataset was curated by STA, ML, and DT. The ExtendedQA dataset was curated by RS and DT. The human analysis was performed by LA. KB, RS, LA, DF, CK, JNK, SN, and DT provided clinical expertise. STA, ML, KB, JNK, and DT provided technical expertise. All authors read the manuscript, contributed to editing, and agreed to the submission of this paper.

*Competing Interests*

ML is employed by Generali Deutschland Services GmbH. JNK declares consulting services for Bioptimus, France; Owkin, France; DoMore Diagnostics, Norway; Panakeia, UK; AstraZeneca, UK; Scailyte, Switzerland; Mindpeak, Germany; and MultiplexDx, Slovakia. Furthermore, he holds shares in StratifAI GmbH, Germany, has received a research grant by GSK, and has received honoraria by AstraZeneca, Bayer, Eisai, Janssen, MSD, BMS, Roche, Pfizer and Fresenius. DT holds shares in StratifAI GmbH and received honoraria for lectures by Bayer. The other authors do not have any competing interests to disclose.

# Supplementary Information

**Table S1: Accuracy results for individual specialties on the RSNA-RadioQA dataset.** Mean accuracy, represented in percent, on RSNA-RadioQA with n=80 questions (details in **Appendix S1**).

| Subspecialty | GPT-3.5-turbo | | GPT-4 | | Mistral-7B | | Mixtral-8x7B | | Llama3-8B | | Llama3-70B | | *Average* | |
|---|---|---|---|---|---|---|---|---|---|---|---|---|---|---|
| | Radio RAG | No-RAG | Radio RAG | No-RAG | Radio RAG | No-RAG | Radio RAG | No-RAG | Radio RAG | No-RAG | Radio RAG | No-RAG | Radio RAG | No-RAG |
| Breast Imaging [n=10] | 90 | 90 | 100 | 100 | 70 | 70 | 80 | 80 | 70 | 70 | 90 | 80 | *83* | *67* |
| Cardiac [n=8] | 50 | 50 | 62 | 62 | 38 | 50 | 62 | 50 | 38 | 50 | 50 | 12 | *50* | *33* |
| Chest [n=13] | 38 | 31 | 46 | 31 | 31 | 31 | 62 | 46 | 31 | 31 | 31 | 31 | *33* | *33* |
| CT [n=28] | 68 | 57 | 79 | 71 | 64 | 54 | 82 | 68 | 57 | 57 | 71 | 68 | *67* | *50* |
| Emergency Radiology [n=6] | 67 | 50 | 67 | 67 | 67 | 33 | 50 | 33 | 33 | 50 | 67 | 67 | *50* | *33* |
| Gastrointestinal [n=12] | 67 | 67 | 83 | 83 | 58 | 42 | 83 | 67 | 58 | 50 | 67 | 58 | *67* | *50* |
| Genitourinary [n=8] | 75 | 88 | 88 | 88 | 88 | 62 | 75 | 75 | 75 | 75 | 75 | 88 | *67* | *67* |
| Head and Neck [n=9] | 67 | 67 | 89 | 89 | 56 | 67 | 67 | 67 | 89 | 44 | 56 | 67 | *67* | *50* |
| MRI [n=20] | 95 | 90 | 90 | 85 | 65 | 70 | 85 | 75 | 60 | 75 | 80 | 80 | *67* | *67* |
| Molecular Imaging [n=11] | 55 | 55 | 64 | 64 | 55 | 55 | 64 | 64 | 55 | 64 | 55 | 45 | *50* | *50* |
| Musculoskeletal [n=14] | 93 | 93 | 93 | 86 | 64 | 71 | 79 | 64 | 64 | 71 | 86 | 86 | *67* | *67* |
| Neuroradiology [n=11] | 73 | 64 | 64 | 73 | 45 | 55 | 82 | 55 | 45 | 36 | 64 | 73 | *50* | *50* |
| Nuclear Medicine [n=13] | 62 | 62 | 69 | 69 | 54 | 54 | 69 | 69 | 62 | 62 | 54 | 46 | *50* | *50* |
| Oncologic Imaging [n=16] | 75 | 69 | 81 | 75 | 69 | 69 | 81 | 75 | 69 | 62 | 69 | 56 | *67* | *67* |
| Pediatric [n=7] | 43 | 14 | 43 | 43 | 29 | 14 | 57 | 14 | 29 | 14 | 43 | 29 | *33* | *17* |
| Radiation Oncology [n=9] | 89 | 89 | 89 | 78 | 78 | 78 | 89 | 89 | 78 | 56 | 78 | 89 | *83* | *67* |
| Ultrasound [n=10] | 90 | 90 | 90 | 90 | 60 | 70 | 80 | 70 | 70 | 70 | 90 | 70 | *67* | *67* |
| Vascular Imaging [n=13] | 77 | 54 | 69 | 69 | 46 | 46 | 69 | 69 | 46 | 62 | 62 | 54 | *50* | *50* |

Table S2: **Overview of hallucination rates across LLMs.** "Context relevant" refers to the percentage of cases where the retrieved articles and context were fully relevant to the question. Cases where the context was relevant but the response was still incorrect are classified as hallucinations. Results are presented for both the RSNA-RadioQA dataset (n=80 questions) and the ExtendedQA dataset (n=24 questions).

|  | RAG+GPT-3.5-turbo | RAG+GPT-4 | RAG+Mistral-7B | RAG+Mixtral-8x7B | RAG+Llama 3-8B | RAG+Llama 3-70B |
|---|---|---|---|---|---|---|
| **RSNA-RadioQA Dataset (n=80)** | | | | | | |
| Context relevant | 72% (58/80) | 72% (58/80) | 72% (58/80) | 72% (58/80) | 72% (58/80) | 72% (58/80) |
| Context relevant, response incorrect (hallucination) | 10% (8/80) | 6% (5/80) | 25% (20/80) | 9% (7/80) | 22% (18/80) | 14% (11/80) |
| Context irrelevant, response correct | 10% (9/80) | 12% (10/80) | 7% (6/80) | 12% (10/80) | 9% (7/80) | 10% (8/80) |
| Context irrelevant, response wrong | 16% (13/80) | 15% (12/80) | 20% (16/80) | 15% (12/80) | 19% (15/80) | 17% (14/80) |
| **ExtendedQA Dataset (n=24)** | | | | | | |
| Context relevant | 83% (20/24) | 83% (20/24) | 83% (20/24) | 83% (20/24) | 83% (20/24) | 83% (20/24) |
| Context relevant, response incorrect (hallucination) | 18% (4/24) | 12% (3/24) | 21% (5/24) | 8% (2/24) | 21% (5/24) | 18% (4/24) |
| Context irrelevant, response correct | 4% (1/24) | 4% (1/24) | 8% (2/24) | 4% (1/24) | 4% (1/24) | 8% (2/24) |
| Context irrelevant, response wrong | 12% (3/24) | 12% (3/24) | 8% (2/24) | 12% (3/24) | 12% (3/24) | 8% (2/24) |

**Table S3**: Answers of the RadioRAG-powered GPT-3.5-turbo to the first twnty questions of the RSNA-RadioQA dataset, together with the provided sources by the platform.

| Question Identifier | LLM's (RadioRAG+GPT-3.5-turbo) Answer | Most Relevant Sources |
|---|---|---|
| RSNA-RadioQA-Q1 | The most likely diagnosis is a benign fat-containing mass with rim calcifications. | https://radiopaedia.org/articles/breast-mass?lang=us, https://radiopaedia.org/articles/breast-imaging-reporting-and-data-system-bi-rads?lang=us |
| RSNA-RadioQA-Q2 | The most likely diagnosis is phyllodes tumor of the breast. | https://radiopaedia.org/articles/breast-mri?lang=us, https://radiopaedia.org/articles/inflammatory-carcinoma-of-the-breast?lang=us |
| RSNA-RadioQA-Q3 | Cat scratch disease lymphadenitis. | https://radiopaedia.org/articles/tuberculous-mastitis?lang=us, https://radiopaedia.org/articles/inflammatory-carcinoma-of-the-breast?lang=us, https://radiopaedia.org/articles/steatocystoma-multiplex-of-the-breast-1?lang=us |
| RSNA-RadioQA-Q4 | The most likely diagnosis is invasive ductal carcinoma in both breasts. | https://radiopaedia.org/articles/bsbr-breast-imaging-classification?lang=us, https://radiopaedia.org/articles/breast-cancer-summary?lang=us, https://radiopaedia.org/articles/triple-receptor-negative-breast-cancer?lang=us |
| RSNA-RadioQA-Q5 | The most likely diagnosis is bronchogenic carcinoma. | https://radiopaedia.org/articles/left-upper-lobe-collapse?lang=us, https://radiopaedia.org/articles/left-upper-lobe-collapse-in-the-exam?lang=us, https://radiopaedia.org/articles/bronchial-anthracofibrosis?lang=us |
| RSNA-RadioQA-Q6 | The most likely diagnosis is organizing pneumonia. | https://radiopaedia.org/articles/acute-eosinophilic-pneumonia?lang=us, https://radiopaedia.org/articles/congestive-cardiac-failure?lang=us, https://radiopaedia.org/articles/rituximab-induced-interstitial-lung-disease-1?lang=us |
| RSNA-RadioQA-Q7 | The most likely diagnosis is Tuberous Sclerosis Complex (TSC). | https://radiopaedia.org/articles/paediatric-cystic-renal-diseases?lang=us, https://radiopaedia.org/articles/tuberous-sclerosis?lang=us |
| RSNA-RadioQA-Q8 | The most likely diagnosis is a patellar fracture. | https://radiopaedia.org/articles/osteoarthritis-of-the-hip?lang=us, https://radiopaedia.org/articles/osteoarthritis-of-the-knee?lang=us |
| RSNA-RadioQA-Q9 | The most likely diagnosis is a post-traumatic aorto-left renal vein fistula. | https://radiopaedia.org/articles/aorto-left-renal-vein-fistula?lang=us, https://radiopaedia.org/articles/pelvic-lipomatosis?lang=us |
| RSNA-RadioQA-Q10 | The most likely diagnosis is Hashimoto's thyroiditis. | https://radiopaedia.org/articles/steatocystoma-multiplex-of-the-breast-1?lang=us, https://radiopaedia.org/articles/acute-non-traumatic-abdominal-pain-in-pregnancy?lang=us, https://radiopaedia.org/articles/non-mass-enhancement-breast-mri?lang=us |
| RSNA-RadioQA-Q11 | The most likely diagnosis is sarcoidosis. | https://radiopaedia.org/articles/dacryoadenitis?lang=us, https://radiopaedia.org/articles/idiopathic-orbital-inflammation?lang=us |
| RSNA-RadioQA-Q12 | The most likely diagnosis is epidural angiolipoma. | https://radiopaedia.org/articles/cauda-equina-syndrome?lang=us, https://radiopaedia.org/articles/extradural-spinal-cavernous-malformation?lang=us, https://radiopaedia.org/articles/low-back-pain?lang=us |
| RSNA-RadioQA-Q13 | The most likely diagnosis is a femoral diaphyseal stress injury. | https://radiopaedia.org/articles/femoral-diaphyseal-stress-injury?lang=us, https://radiopaedia.org/articles/osteoarthritis-of-the-hip?lang=us |
| RSNA-RadioQA-Q14 | The most likely diagnosis is Kienbock's disease. | https://radiopaedia.org/articles/ulnar-impaction-syndrome?lang=us, https://radiopaedia.org/articles/kienbock-disease-2?lang=us |
| RSNA-RadioQA-Q15 | LR-1 - benign simple cyst. | https://radiopaedia.org/articles/li-rads-us-surveillance?lang=us, https://radiopaedia.org/articles/ctmri-li-rads?lang=us |
| RSNA-RadioQA-Q16 | The most likely diagnosis is Tenosynovial giant cell tumor (TGCT). | https://radiopaedia.org/articles/soft-tissue-mass?lang=us, https://radiopaedia.org/articles/tenosynovial-giant-cell-tumour-2?lang=us, https://radiopaedia.org/articles/flexor-pronator-mass-injury?lang=us |
| RSNA-RadioQA-Q17 | The most likely diagnosis is alveolar soft part sarcoma (ASPS). | https://radiopaedia.org/articles/alveolar-soft-part-sarcoma-2?lang=us, https://radiopaedia.org/articles/stromal-tumour-of-uncertain-malignant-potential?lang=us |
| RSNA-RadioQA-Q18 | The most likely diagnosis is an old avulsion fracture of the inferior pole of the patella with remodeling. | https://radiopaedia.org/articles/anterior-cruciate-ligament-avulsion-fracture?lang=us, https://radiopaedia.org/articles/tibial-tuberosity-avulsion-fracture?lang=us, https://radiopaedia.org/articles/paediatric-knee-radiograph-an-approach?lang=us |
| RSNA-RadioQA-Q19 | The most likely diagnosis is a benign complicated cyst in the right breast. | https://radiopaedia.org/articles/mature-cystic-ovarian-teratoma-1?lang=us, https://radiopaedia.org/articles/nodular-pulmonary-amyloidosis?lang=us, https://radiopaedia.org/articles/mammary-duct-ectasia?lang=us |
| RSNA-RadioQA-Q20 | The most likely diagnosis is a retroperitoneal teratoma. | https://radiopaedia.org/articles/peritoneal-inclusion-cyst?lang=us |

**Table S4: Answers of the conventional GPT-3.5-turbo to the first thirty questions of RSNA-RadioQA dataset.** For the original questions and their ground truth answers of the RSNA-RadioQA dataset, refer to **Appendix S1.**

| Question Identifier | LLM's (GPT-3.5-turbo) Answer |
| --- | --- |
| RSNA-RadioQA-Q1 | The most likely diagnosis is a benign breast lesion, such as a lipoma or fibroadenoma, with associated calcifications. |
| RSNA-RadioQA-Q2 | The most likely diagnosis is desmoid tumor (also known as aggressive fibromatosis). |
| RSNA-RadioQA-Q3 | The most likely diagnosis is Cat Scratch Disease (CSD) with suppurative granulomatous lymphadenitis. |
| RSNA-RadioQA-Q4 | The most likely diagnosis is bilateral invasive ductal carcinoma of the breast, ER positive, PR positive, and HER2 negative, with no lymph node involvement. |
| RSNA-RadioQA-Q5 | The most likely diagnosis is pulmonary rhabdomyosarcoma. |
| RSNA-RadioQA-Q6 | The most likely diagnosis is severe COVID-19 pneumonia. |
| RSNA-RadioQA-Q7 | The most likely diagnosis is pulmonary lymphangioleiomyomatosis (LAM), a characteristic manifestation of Tuberous Sclerosis Complex (TSC). |
| RSNA-RadioQA-Q8 | The most likely diagnosis is a patellar fracture. |
| RSNA-RadioQA-Q9 | The most likely diagnosis is diverticulitis with associated pericolic inflammation and phlegmon formation. |
| RSNA-RadioQA-Q10 | The most likely diagnosis is Hashimoto's thyroiditis. |
| RSNA-RadioQA-Q11 | The most likely diagnosis is orbital sarcoidosis. |
| RSNA-RadioQA-Q12 | The most likely diagnosis is epidural spinal hemangioma. |
| RSNA-RadioQA-Q13 | The most likely diagnosis is a stress fracture of the left femoral neck. |
| RSNA-RadioQA-Q14 | The most likely diagnosis is Kienböck's disease. |
| RSNA-RadioQA-Q15 | The most likely diagnosis is a benign simple cyst of the liver. |
| RSNA-RadioQA-Q16 | The most likely diagnosis is Giant Cell Tumor of Tendon Sheath (GCTTS). |
| RSNA-RadioQA-Q17 | The most likely diagnosis is Alveolar soft part sarcoma (ASPS). |
| RSNA-RadioQA-Q18 | The most likely diagnosis is patellar tendon avulsion fracture. |
| RSNA-RadioQA-Q19 | The most likely diagnosis is benign complicated cysts, given the resolution of the mass with aspiration and the benign nature of the identified cysts on imaging. |
| RSNA-RadioQA-Q20 | The most likely diagnosis is a retroperitoneal teratoma. |
| RSNA-RadioQA-Q21 | The most likely diagnosis is solitary fibrous tumor (SFT) of the thigh. |
| RSNA-RadioQA-Q22 | The most likely diagnosis is a renal artery aneurysm. |
| RSNA-RadioQA-Q23 | The most likely diagnosis is popliteal artery entrapment syndrome. |
| RSNA-RadioQA-Q24 | The most likely diagnosis is osteosarcoma. |
| RSNA-RadioQA-Q25 | The most likely diagnosis is pulmonary embolism. |
| RSNA-RadioQA-Q26 | The most likely diagnosis is recurrent pericarditis, possibly related to systemic lupus erythematosus. |
| RSNA-RadioQA-Q27 | The most likely diagnosis is gastroschisis with associated omphalocele and cleft lip/palate. |
| RSNA-RadioQA-Q28 | The most likely diagnosis is adenocarcinoma of the ileocecal region. |
| RSNA-RadioQA-Q29 | Biliary atresia. |
| RSNA-RadioQA-Q30 | The most likely diagnosis is an incarcerated inguinal hernia with an acutely inflamed, herniated appendix, known as Amyand's hernia. |

# Appendix S1

The Radiological Society of North America Case Collection adapted to RadioRAG pipeline question answering (RSNA-RadioQA) dataset, consists of 80 expert-curated questions. The complete dataset with all the questions and their corresponding answers is as follows.

**Question Identifier:** RSNA-RadioQA-Q1

**DOI of the Original Case:** 10.1148/cases.20227914.
**Authors of the Original Case:** K. Elzinga, R. Woods.
**Title of the Original Case**: BI-RADS 2: Rim Calcifications.
**Publication Date of the Original Case:** 10/10/2022.
**Supspecialties:** Breast Imaging.

**Question Identifier:** RSNA-RadioQA-Q2

**DOI of the Original Case:** 10.1148/cases.20227478.
**Authors of the Original Case:** A. Aripoli, P. Iglar, E. Friedman.
**Title of the Original Case**: Breast Fibromatosis.
**Publication Date of the Original Case:** 8/8/2022.
**Supspecialties:** Breast Imaging.

**Question Identifier:** RSNA-RadioQA-Q3

**DOI of the Original Case:** 10.1148/cases.20239154.
**Authors of the Original Case:** C. Ayeni, L. Misbach, M. Quintana, P. Slanetz.
**Title of the Original Case**: Cat-Scratch Disease.
**Publication Date of the Original Case:** 4/25/2023.
**Supspecialties:** Breast Imaging, Ultrasound.

**Question Identifier:** RSNA-RadioQA-Q4

**DOI of the Original Case:** 10.1148/cases.20238930.
**Authors of the Original Case:** L. Shah, T. Kuritza, S. Benjamin, R. Ganesh.
**Title of the Original Case**: Bilateral Synchronous Invasive Ductal Carcinoma in a Male.
**Publication Date of the Original Case:** 3/17/2023.
**Supspecialties:** Breast Imaging, Ultrasound.

**Question Identifier:** RSNA-RadioQA-Q5

**DOI of the Original Case:** 10.1148/cases.20239212.
**Authors of the Original Case:** C. Walker, S. Gerrie, M. Aquino.
**Title of the Original Case**: Pleuropulmonary Blastoma.
**Publication Date of the Original Case:** 5/19/2023.
**Supspecialties:** Chest, Pediatric.

**Question Identifier:** RSNA-RadioQA-Q6

**DOI of the Original Case:** 10.1148/cases.20238747.
**Authors of the Original Case:** J. Daniel, S. Ayad Al-Katib.
**Title of the Original Case**: Post-COVID Interstitial Lung Disease.
**Publication Date of the Original Case:** 3/3/2023.
**Supspecialties:** Computed Tomography, Chest.

**Question Identifier:** RSNA-RadioQA-Q7

**DOI of the Original Case:** 10.1148/cases.20238883.
**Authors of the Original Case:** N. Raval, R. Jha, P. Bergquist, N. Jain.

**Title of the Original Case**: Multifocal Micronodular Pneumocyte Hyperplasia.
**Publication Date of the Original Case:** 2/6/2023.
**Supspecialties:** Computed Tomography, Chest.

**Question Identifier:** RSNA-RADIOQA-Q8

**DOI of the Original Case:** 10.1148/cases.20227185.
**Authors of the Original Case:** B. Guthridge, B. Fink.
**Title of the Original Case**: Vertical Patellar Fracture.
**Publication Date of the Original Case:** 12/1/2022.
**Supspecialties:** Emergency Radiology, Musculoskeletal.

**Question Identifier:** RSNA-RadioQA-Q9

**DOI of the Original Case:** 10.1148/cases.20226345.
**Authors of the Original Case:** B. Tallman, R. Jarman.
**Title of the Original Case**: Epiploic Appendagitis.
**Publication Date of the Original Case:** 1/31/2022.
**Supspecialties:** Gastrointestinal, Emergency Radiology, Computed Tomography.

**Question Identifier:** RSNA-RadioQA-Q10

**DOI of the Original Case:** 10.1148/cases.20224821.
**Authors of the Original Case:** Q. Li, J. Wang, D. Gao, T. Pierce.
**Title of the Original Case**: Hashimoto's thyroiditis.
**Publication Date of the Original Case:** 4/4/2022.
**Supspecialties:** Ultrasound, Head and Neck.

**Question Identifier:** RSNA-RadioQA-Q11

**DOI of the Original Case:** 10.1148/cases.20238331.
**Authors of the Original Case:** A. Kumar, D. Gewolb, A. Narayan.
**Title of the Original Case**: Orbital Sarcoidosis.
**Publication Date of the Original Case:** 2/15/2023.
**Supspecialties:** Head and Neck, Neuroradiology.

**Question Identifier:** RSNA-RadioQA-Q12

**DOI of the Original Case:** 10.1148/cases.20224694.
**Authors of the Original Case:** L. Chiu, J. Yoon
**Title of the Original Case**: Spinal Angiolipoma.
**Publication Date of the Original Case:** 11/10/2022.
**Supspecialties:** Neuroradiology, Magnetic Resonance Imaging.

**Question Identifier:** RSNA-RadioQA-Q13

**DOI of the Original Case:** 10.1148/cases.20227653.
**Authors of the Original Case:** G. Rahmani.
**Title of the Original Case**: Femoral neck stress fracture.
**Publication Date of the Original Case:** 12/8/2022.
**Supspecialties:** Magnetic Resonance Imaging, Musculoskeletal.

**Question Identifier:** RSNA-RadioQA-Q14

**DOI of the Original Case:** 10.1148/cases.20227689.
**Authors of the Original Case:** T. Schermann, R. Potenza, T. DenOtter.
**Title of the Original Case**: Kienbock Disease.
**Publication Date of the Original Case:** 12/19/2022.
**Supspecialties:** Musculoskeletal, Magnetic Resonance Imaging.

**Question Identifier:** RSNA-RadioQA-Q15

**DOI of the Original Case:** 10.1148/cases.20238060.
**Authors of the Original Case:** A. Alkhudari, A. Gibson, J. Lee, A. Sobieh.
**Title of the Original Case**: LI-RADS 1.
**Publication Date of the Original Case:** 3/6/2023.
**Supspecialties:** Gastrointestinal, Magnetic Resonance Imaging.
Copyright © Radiological Society of North America, Inc. (RSNA), All Rights Reserved

**Question Identifier:** RSNA-RadioQA-Q16

**DOI of the Original Case:** 10.1148/cases.20238375.
**Authors of the Original Case:** J. Paek, R. Rozzi, J. Judge.
**Title of the Original Case**: Tenosynovial Giant Cell Tumor of the Finger.
**Publication Date of the Original Case:** 3/17/2023.
**Supspecialties:** Musculoskeletal, Magnetic Resonance Imaging.
Copyright © Radiological Society of North America, Inc. (RSNA), All Rights Reserved

**Question Identifier:** RSNA-RadioQA-Q17

**DOI of the Original Case:** 10.1148/cases.20237558.
**Authors of the Original Case:** B. Franz, P. Patel, C. Scher.
**Title of the Original Case**: Alveolar Soft Part Sarcoma.
**Publication Date of the Original Case:** 4/18/2023.
**Supspecialties:** Musculoskeletal, Magnetic Resonance Imaging.
Copyright © Radiological Society of North America, Inc. (RSNA), All Rights Reserved

**Question Identifier:** RSNA-RadioQA-Q18

**DOI of the Original Case:** 10.1148/cases.20238620.
**Authors of the Original Case:** R. Iyer, M. Kumaravel.
**Title of the Original Case**: Patellar tendon tear.
**Publication Date of the Original Case:** 5/12/2023.
**Supspecialties:** Musculoskeletal, Magnetic Resonance Imaging.
Copyright © Radiological Society of North America, Inc. (RSNA), All Rights Reserved

**Question Identifier:** RSNA-RadioQA-Q19

**DOI of the Original Case:** 10.1148/cases.20225780.
**Authors of the Original Case:** N. Vu, R. Woods.
**Title of the Original Case**: Complicated Breast Cyst.
**Publication Date of the Original Case:** 2/24/2022.
**Supspecialties:** Magnetic Resonance Imaging, Breast Imaging.
Copyright © Radiological Society of North America, Inc. (RSNA), All Rights Reserved

**Question Identifier:** RSNA-RadioQA-Q20

**DOI of the Original Case:** 10.1148/cases.20238293.
**Authors of the Original Case:** S. Carter, F. Flaherty.
**Title of the Original Case**: Primary Retroperitoneal Mature Cystic Teratoma.
**Publication Date of the Original Case:** 3/27/2023.
**Supspecialties:** Magnetic Resonance Imaging, Computed Tomography.
Copyright © Radiological Society of North America, Inc. (RSNA), All Rights Reserved

**Question Identifier:** RSNA-RadioQA-Q21

**DOI of the Original Case:** 10.1148/cases.20237602.
**Authors of the Original Case:** L. Verst, D. Constantino, M. Chalian.
**Title of the Original Case**: Solitary Fibrous Tumor.
**Publication Date of the Original Case:** 2/6/2023.
**Supspecialties:** Musculoskeletal, Magnetic Resonance Imaging, Ultrasound.
Copyright © Radiological Society of North America, Inc. (RSNA), All Rights Reserved

**Question Identifier:** RSNA-RadioQA-Q22

**DOI of the Original Case:** 10.1148/cases.20238276.
**Authors of the Original Case:** S. Goddard, A. Annamalai, C. Chamberlin, B. Triche.
**Title of the Original Case**: Renal Arteriovenous Malformation.
**Publication Date of the Original Case:** 6/8/2023.
**Supspecialties:** Vascular Imaging, Interventional Radiology, Computed Tomography, Genitourinary.
Copyright © Radiological Society of North America, Inc. (RSNA), All Rights Reserved

**DOI of the Original Case:** 10.1148/cases.20238491.
**Authors of the Original Case:** E. Berger, M. MacDonald.
**Title of the Original Case**: Popliteal Artery Entrapment Syndrome with Thrombosis.
**Publication Date of the Original Case:** 6/8/2023.
**Supspecialties:** Ultrasound, Computed Tomography, Vascular Imaging, Musculoskeletal.
Copyright © Radiological Society of North America, Inc. (RSNA), All Rights Reserved

**Question Identifier:** RSNA-RadioQA-Q24

**DOI of the Original Case:** 10.1148/cases.20238762.
**Authors of the Original Case:** N. LeCrone, A. Goggins, Y. Qiao, A. Salem.
**Title of the Original Case**: Conventional Osteosarcoma.
**Publication Date of the Original Case:** 6/7/2023.
**Supspecialties:** Musculoskeletal.
Copyright © Radiological Society of North America, Inc. (RSNA), All Rights Reserved

**Question Identifier:** RSNA-RadioQA-Q25

**DOI of the Original Case:** 10.1148/cases.20239055.
**Authors of the Original Case:** F. Lo, S. Robert, G. Brahm.
**Title of the Original Case**: Epipericardial Fat Necrosis.
**Publication Date of the Original Case:** 3/3/2023.
**Supspecialties:** Computed Tomography, Cardiac, Emergency Radiology, Chest.
Copyright © Radiological Society of North America, Inc. (RSNA), All Rights Reserved

**Question Identifier:** RSNA-RadioQA-Q26

**DOI of the Original Case:** 10.1148/cases.20226457.
**Authors of the Original Case:** A. Canan, N. Cabrera.
**Title of the Original Case**: Cardiac tamponade.
**Publication Date of the Original Case:** 1/16/2022.
**Supspecialties:** Magnetic Resonance Imaging, Computed Tomography, Cardiac.
Copyright © Radiological Society of North America, Inc. (RSNA), All Rights Reserved

**Question Identifier:** RSNA-RadioQA-Q27

**DOI of the Original Case:** 10.1148/cases.20239046.
**Authors of the Original Case:** V. Krishnan, S. Jaganathan, K. Schmitz, M. Renno.
**Title of the Original Case**: Pentalogy of Cantrell.
**Publication Date of the Original Case:** 3/3/2023.
**Supspecialties:** Cardiac, Gastrointestinal, Computed Tomography, Pediatric.
Copyright © Radiological Society of North America, Inc. (RSNA), All Rights Reserved

**Question Identifier:** RSNA-RadioQA-Q28

**DOI of the Original Case:** 10.1148/cases.20227592.
**Authors of the Original Case:** Y. Park, O. Kalinkin.
**Title of the Original Case**: Small Bowel Carcinoid Tumor.
**Publication Date of the Original Case:** 7/11/2022.
**Supspecialties:** Gastrointestinal, Computed Tomography.
Copyright © Radiological Society of North America, Inc. (RSNA), All Rights Reserved

**Question Identifier:** RSNA-RadioQA-Q29

**DOI of the Original Case:** 10.1148/cases.20228374.
**Authors of the Original Case:** K. Banks.
**Title of the Original Case**: Biliary Atresia.
**Publication Date of the Original Case:** 10/20/2022.
**Supspecialties:** Nuclear Medicine, Gastrointestinal.
Copyright © Radiological Society of North America, Inc. (RSNA), All Rights Reserved

**Question Identifier:** RSNA-RadioQA-Q30

**DOI of the Original Case:** 10.1148/cases.20227840.
**Authors of the Original Case:** C. Qian, N. Parikh, J. Oh, J. Amorosa.
**Title of the Original Case**: Amyand Hernia with Appendicitis.
**Publication Date of the Original Case:** 11/2/2022.
**Supspecialties:** Genitourinary, Gastrointestinal, Computed Tomography.

**Question Identifier:** RSNA-RadioQA-Q31

**DOI of the Original Case:** 10.1148/cases.20225796.
**Authors of the Original Case:** A. Bamashmos, K. Elfatairy, R. Hegde, O. Awan.
**Title of the Original Case**: Cervical spine gout.
**Publication Date of the Original Case:** 1/14/2022.
**Supspecialties:** Computed Tomography, Musculoskeletal.

**Question Identifier:** RSNA-RadioQA-Q32

**DOI of the Original Case:** 10.1148/cases.20227086.
**Authors of the Original Case:** J. Benjamin, H. Son.
**Title of the Original Case**: Septic Embolism.
**Publication Date of the Original Case:** 4/15/2022.
**Supspecialties:** Chest, Computed Tomography.

**Question Identifier:** RSNA-RadioQA-Q33

**DOI of the Original Case:** 10.1148/cases.20227539.
**Authors of the Original Case:** A. Shah, O. Shah, T. Shera, S. Shabir.
**Title of the Original Case**: Erdheim-Chester disease.
**Publication Date of the Original Case:** 9/21/2022.
**Supspecialties:** Computed Tomography, Chest.

**Question Identifier:** RSNA-RadioQA-Q34

**DOI of the Original Case:** 10.1148/cases.20226333.
**Authors of the Original Case:** Z. Timmerman, M. Carrillo, B. Shah.
**Title of the Original Case**: Male invasive Ductal Carcinoma.
**Publication Date of the Original Case:** 1/19/2022.
**Supspecialties:** Ultrasound, Breast Imaging, Oncologic Imaging.

**Question Identifier:** RSNA-RadioQA-Q35

**DOI of the Original Case:** 10.1148/cases.20239100.
**Authors of the Original Case:** J. Eichhorn, M. Fox.
**Title of the Original Case**: FDG -avid axillary lymphadenopathy.
**Publication Date of the Original Case:** 2/15/2023.
**Supspecialties:** Nuclear Medicine, Oncologic Imaging, Molecular Imaging, Chest.

**Question Identifier:** RSNA-RadioQA-Q36

**DOI of the Original Case:** 10.1148/cases.20239042.
**Authors of the Original Case:** J. Eichhorn, N. Phelan, J. Gilstrap.
**Title of the Original Case**: Polyostotic Paget Disease.
**Publication Date of the Original Case:** 3/3/2023.
**Supspecialties:** Musculoskeletal, Molecular Imaging, Breast Imaging, Nuclear Medicine, Oncologic Imaging.

**Question Identifier:** RSNA-RadioQA-Q37

**DOI of the Original Case:** 10.1148/cases.20238698.
**Authors of the Original Case:** D. Mehta, S. Shinde.
**Title of the Original Case**: Undifferentiated Embryonal Sarcoma.
**Publication Date of the Original Case:** 1/17/2023.
**Supspecialties:** Pediatric, Oncologic Imaging.

**Question Identifier:** RSNA-RadioQA-Q38

**DOI of the Original Case:** 10.1148/cases.20228072.
**Authors of the Original Case:** I. Buren, A. Fung.
**Title of the Original Case**: LI-RADS TIV.

## Question Identifier: RSNA-RadioQA-Q39

## Question Identifier: RSNA-RadioQA-Q40

## Question Identifier: RSNA-RadioQA-Q41

## Question Identifier: RSNA-RadioQA-Q42

## Question Identifier: RSNA-RadioQA-Q43

## Question Identifier: RSNA-RadioQA-Q45

## Question Identifier: RSNA-RadioQA-Q46

**Publication Date of the Original Case:** 11/4/2020.
**Supspecialties:** Pediatric, Cardiac, Magnetic Resonance Imaging, Ultrasound, Obstetrics/Gynecology, Neuroradiology.

**Question Identifier:** RSNA-RadioQA-Q55

**DOI of the Original Case:** 10.1148/cases.20223038.
**Authors of the Original Case:** E. Barkovich, N. Khati.
**Title of the Original Case**: Subclavian Stenosis with Pre-Steal.
**Publication Date of the Original Case:** 9/8/2021.
**Supspecialties:** Vascular Imaging, Ultrasound.

**Question Identifier:** RSNA-RadioQA-Q56

**DOI of the Original Case:** 10.1148/cases.20225859.
**Authors of the Original Case:** J. Madsen, C. Cooley, C. Dumas.
**Title of the Original Case**: Aortoiliac Occlusive Disease.
**Publication Date of the Original Case:** 9/29/2021.
**Supspecialties:** Computed Tomography, Vascular Imaging.

**Question Identifier:** RSNA-RadioQA-Q57

**DOI of the Original Case:** 10.1148/cases.20224513.
**Authors of the Original Case:** D. Bittles, A. Gunn.
**Title of the Original Case**: Post-Traumatic Arteriovenous Fistula.
**Publication Date of the Original Case:** 4/21/2021.
**Supspecialties:** Interventional Radiology, Vascular Imaging.

**Question Identifier:** RSNA-RadioQA-Q58

**DOI of the Original Case:** 10.1148/cases.20224581.
**Authors of the Original Case:** J. Liu, V. Gorolay.
**Title of the Original Case**: Acute Aortoiliac Occlusion.
**Publication Date of the Original Case:** 3/15/2021.
**Supspecialties:** Computed Tomography, Vascular Imaging.

**Question Identifier:** RSNA-RadioQA-Q59

**DOI of the Original Case:** 10.1148/cases.20222067.
**Authors of the Original Case:** J. Shen, R. Lim.
**Title of the Original Case**: Pulmonary arteriovenous malformation.
**Publication Date of the Original Case:** 6/16/2020.
**Supspecialties:** Cardiac, Vascular Imaging, Chest.

**Question Identifier:** RSNA-RadioQA-Q60

**DOI of the Original Case:** 10.1148/cases.20222080.
**Authors of the Original Case:** J. Shen, J. Tse, R. Lim.
**Title of the Original Case**: Pulmonary Embolism with Right Heart Strain.
**Publication Date of the Original Case:** 5/15/2020.
**Supspecialties:** Cardiac, Chest, Vascular Imaging.

**Question Identifier:** RSNA-RadioQA-Q61

**DOI of the Original Case:** 10.1148/cases.20223188.
**Authors of the Original Case:** O. Yusufzai, P. Patel.
**Title of the Original Case**: Thoracic Aortic Pseudoaneurysm.
**Publication Date of the Original Case:** 4/6/2021.
**Supspecialties:** Chest, Vascular Imaging, Computed Tomography.

**Question Identifier:** RSNA-RadioQA-Q62

**DOI of the Original Case:** 10.1148/cases.20223594.
**Authors of the Original Case:** A. Ritchey, J. Kucera.
**Title of the Original Case**: Takayasu Arteritis.
**Publication Date of the Original Case:** 12/7/2020.
**Supspecialties:** Interventional Radiology, Pediatric, Vascular Imaging.

**Question Identifier:** RSNA-RadioQA-Q63

**DOI of the Original Case:** 10.1148/cases.20223946.
**Authors of the Original Case:** R. Davidyan, S. Singh.
**Title of the Original Case**: Duplicated Inferior Vena Cava.
**Publication Date of the Original Case:** 9/15/2021.
**Supspecialties:** Computed Tomography, Vascular Imaging, Interventional Radiology.

**Question Identifier:** RSNA-RadioQA-Q64

**DOI of the Original Case:** 10.1148/cases.20224487.
**Authors of the Original Case:** S. Srinivas, T. Retson.
**Title of the Original Case**: Cerebral Arteriovenous Malformation.
**Publication Date of the Original Case:** 5/27/2021.
**Supspecialties:** Neuroradiology, Vascular Imaging, Radiation Oncology, Magnetic Resonance Imaging.

**Question Identifier:** RSNA-RadioQA-Q65

**DOI of the Original Case:** 10.1148/cases.20224616.
**Authors of the Original Case:** D. Albenda, M. Queiroz.
**Title of the Original Case**: Encapsulated fat necrosis.
**Publication Date of the Original Case:** 6/15/2021.
**Supspecialties:** Molecular Imaging, Radiation Oncology, Gastrointestinal, Oncologic Imaging, Nuclear Medicine.

**Question Identifier:** RSNA-RadioQA-Q66

**DOI of the Original Case:** 10.1148/cases.20221463.
**Authors of the Original Case:** R. Zeitoun, R. Merard, U. Kularatne.
**Title of the Original Case**: Dermatofibrosarcoma Protuberans.
**Publication Date of the Original Case:** 1/15/2021.
**Supspecialties:** Chest, Radiation Oncology, Computed Tomography, Magnetic Resonance Imaging, Oncologic Imaging, Musculoskeletal.

**Question Identifier:** RSNA-RadioQA-Q67

**DOI of the Original Case:** 10.1148/cases.20238226.
**Authors of the Original Case:** Z. LeBaron, J. Rabang, D. Gridley, A. Roh.
**Title of the Original Case**: Gallbladder Carcinoma.
**Publication Date of the Original Case:** 1/4/2023.
**Supspecialties:** Radiation Oncology, Magnetic Resonance Imaging, Emergency Radiology, Gastrointestinal, Ultrasound, Computed Tomography, Oncologic Imaging.

**Question Identifier:** RSNA-RadioQA-Q68

**DOI of the Original Case:** 10.1148/cases.20224397.
**Authors of the Original Case:** R. Vyas, R. Vyas, A. Singh.
**Title of the Original Case**: Central Nervous System Tuberculosis.
**Publication Date of the Original Case:** 6/7/2021.
**Supspecialties:** Neuroradiology.

**Question Identifier:** RSNA-RadioQA-Q69

**DOI of the Original Case:** 10.1148/cases.20223494.
**Authors of the Original Case:** H. Diep, D. Poletto.
**Title of the Original Case**: Osmotic demyelination syndrome.
**Publication Date of the Original Case:** 11/27/2020.
**Supspecialties:** Neuroradiology.

**Question Identifier:** RSNA-RadioQA-Q70

**DOI of the Original Case:** 10.1148/cases.20224037.
**Authors of the Original Case:** M. Boukoucha, R. Khelifa, A. Daghfous.
**Title of the Original Case**: Leptomeningeal siderosis.
**Publication Date of the Original Case:** 7/7/2021.
**Supspecialties:** Neuroradiology.

**Question Identifier:** RSNA-RadioQA-Q71

.

**DOI of the Original Case:** 10.1148/cases.20224460.
**Authors of the Original Case:** G. Rauch.
**Title of the Original Case**: BI-RADS 2 - Large rod-like calcifications.
**Publication Date of the Original Case:** 7/27/2021.
**Supspecialties:** Breast Imaging.

**Question Identifier:** RSNA-RadioQA-Q72

**DOI of the Original Case:** 10.1148/cases.20225300.
**Authors of the Original Case:** A. Griffith, R. Woods.
**Title of the Original Case**: Benign Breast Lipoma.
**Publication Date of the Original Case:** 7/27/2021.
**Supspecialties:** Breast Imaging.

**Question Identifier:** RSNA-RadioQA-Q73

**DOI of the Original Case:** 10.1148/cases.20225240.
**Authors of the Original Case:** A. Griffith, R. Woods.
**Title of the Original Case**: Axillary Adenopathy Secondary to HIV.
**Publication Date of the Original Case:** 11/1/2021.
**Supspecialties:** Breast Imaging.

**Question Identifier:** RSNA-RadioQA-Q74

**DOI of the Original Case:** 10.1148/cases.20223554.
**Authors of the Original Case:** P. Garg, A. Thakral, R. Malhotra, P. SARGAR.
**Title of the Original Case**: Mineralising vasculopathy with infarct.
**Publication Date of the Original Case:** 1/29/2021.
**Supspecialties:** Head and Neck.

**Question Identifier:** RSNA-RadioQA-Q75

**DOI of the Original Case:** 10.1148/cases.20225940.
**Authors of the Original Case:** A. Velasco, C. Manjarrez, M. Matos.
**Title of the Original Case**: TI-RADS 5.
**Publication Date of the Original Case:** 10/13/2021.
**Supspecialties:** Ultrasound, Head and Neck.

**Question Identifier:** RSNA-RadioQA-Q76

**DOI of the Original Case:** 10.1148/cases.20224673.
**Authors of the Original Case:** T. Albataineh, T. Rizvi.
**Title of the Original Case**: TIPIC syndrome.
**Publication Date of the Original Case:** 6/28/2021.
**Supspecialties:** Head and Neck, Neuroradiology.

**Question Identifier:** RSNA-RadioQA-Q77

**DOI of the Original Case:** 10.1148/cases.20224400.
**Authors of the Original Case:** R. Vyas, A. Vyas, R. Vyas, A. Singh.

**Title of the Original Case**: Right vocal cord palsy.
**Publication Date of the Original Case:** 6/15/2021.
**Supspecialties:** Neuroradiology, Head and Neck.

**Question Identifier:** RSNA-RadioQA-Q78

**DOI of the Original Case:** 10.1148/cases.20225342.
**Authors of the Original Case:** O. Schoeck, M. Horrow, J. Ganeles, J. Gubernick.
**Title of the Original Case**: Putty Kidney.
**Publication Date of the Original Case:** 9/29/2021.
**Supspecialties:** Genitourinary.

**Question Identifier:** RSNA-RadioQA-Q79

**DOI of the Original Case:** 10.1148/cases.20221886.
**Authors of the Original Case:** P. Cormier, K. Ledbetter, J. Leschied, G. Sabharwal.
**Title of the Original Case**: Duplicated Collecting System with Vesicoureteral Reflux in the Lower Pole Moiety.
**Publication Date of the Original Case:** 8/12/2020.
**Supspecialties:** Genitourinary, Pediatric.

**Question Identifier:** RSNA-RadioQA-Q80

**DOI of the Original Case:** 10.1148/cases.20222850.
**Authors of the Original Case:** A. Rand, D. Akselrod, J. Tam.
**Title of the Original Case**: Schistosomiasis Bladder Wall Calcification.
**Publication Date of the Original Case:** 11/2/2020.
**Supspecialties:** Computed Tomography, Genitourinary.

# Appendix S2

The complete ExtendedQA dataset, consisting of 24 expert-curated questions, with all the questions and their corresponding answers is as follows:

| |
|---|
| **Question Identifier:** ExtendedQA-Q1 |

**Question:** A 74 year-old patient presents with progressive shortness of breath over the last years. He is a former mining worker. CT shows calcified pleural plaques. What is the most likely diagnosis?
**Reference Standard Answer:** Asbestosis.
**Supspecialties:** Chest, Computed Tomography.

| |
|---|
| **Question Identifier:** ExtendedQA-Q2 |

**Question:** What is considered a malignant course of the coronary arteries?
**Reference Standard Answer:** Interarterial course.
**Supspecialties:** Cardiac.

| |
|---|
| **Question Identifier:** ExtendedQA-Q3 |

**Question:** What are the differential diagnoses of a solid mass in the anterosuperior mediastinum?
**Reference Standard Answer:** Thymoma, thyroid (mass), lymphoma, teratoma (and germ cell tumors), thoracic aorta (e.g., aneurysm).
**Supspecialties:** Chest.

| |
|---|
| **Question Identifier:** ExtendedQA-Q4 |

**Question:** What are differential diagnoses of pulmonary cavities?
**Reference Standard Answer:** Cancer, autoimmune diseases (granulomatosis with polyangiitis, rheumatoid arthritis), vascular (bland/septic emboli), infection (e.g., tuberculosis), trauma (pneumatoceles), youth (CPAM, sequestration, bronchogenic cyst); Cavity mnemonic.
**Supspecialties:** Chest.

| |
|---|
| **Question Identifier:** ExtendedQA-Q5 |

**Question:** Which MRI contrast agent can be used to detect an insufficiency of a biliary anastomosis?
**Reference Standard Answer:** Eovist / Gadoxetic acid or disodium.
**Supspecialties:** Gastrointestinal, Magnetic Resonance Imaging.

| |
|---|
| **Question Identifier:** ExtendedQA-Q6 |

**Question:** What are possible distributions of pumonary nodules? Name one differential diagnosis for each distribution.
**Reference Standard Answer:** Perilymphatic (e.g., sarcoidosis), centrilobular (special form: tree-in-bud; e.g., tuberculosis), diffuse (e.g., disseminated pulmonary metastasis).
**Supspecialties:** Chest.

| |
|---|
| **Question Identifier:** ExtendedQA-Q7 |

**Question:** I am looking at a prostate MRI and see a lesion in the right posterolateral peripheral zone. The lesion is hypointense in T2, hyperintense in DWI, hypointense in ADC and has a strong and early signal enhancement after contrast administration. What kind of lesion could this be?
**Reference Standard Answer:** Prostate cancer.
**Supspecialties:** Genitourinary, Magnetic Resonance Imaging.

| |
|---|
| **Question Identifier:** ExtendedQA-Q8 |

**Question:** I am looking at an X-ray of the lumbar spine on a sagittal view. There are prominent endplate densities at multiple contiguous vertebral levels, leading to a sclerotic-lucent-sclerotic appearance. The Patient has a history of renal insufficiency. What is this appearance of the spine called and what is the most likely diagnosis?
**Reference Standard Answer:** Rugger jersey spine; secondary hyperparathyroidism.
**Supspecialties:** Musculoskeletal.

| |
|---|
| **Question Identifier:** ExtendedQA-Q9 |

**Question:** I see a prominent pulmonary embolus in the right pulmonaly artery on a CT pulmonary angiography. What are signs of right ventricular dysfunction?
**Reference Standard Answer:** Abnormal position of the interventricular septum, contrast reflux into inferior vena cava/liver veins, increased diameter of the right ventricle.
**Supspecialties:** Cardiac, Chest, Computed Tomography, Emergency Radiology.

| |
|---|
| **Question Identifier:** ExtendedQA-Q10 |

**Question:** A 24 year-old patient presents with right upper quadrant pain, fever and jaundice. MRI of the liver shows T1-hypointense, T2-hyperintense, segmental dilatation of the intrahepatic bile ducts, leading to a multicystic appearance. There are multiple concrements in the intrahepatic bile ducts. The patient has no history of bowel disease. What is the most likely diagnosis?

**Reference Standard Answer:** Caroli-disease.
**Supspecialties:** Emergency Radiology, Musculoskeletal.

**Question Identifier:** ExtendedQA-Q11
**Question:** I am looking at an X-ray of the knee on a anteroposterior view of a patient after trauma. I see an elliptic bone fragment projected parallel to the lateral aspect of the tibial plateau. What is the most likely diagnosis?
**Reference Standard Answer:** Segond fracture, indicative of a rupture of the anterior cruciate ligament.
**Supspecialties:** Gastrointestinal, Magnetic Resonance Imaging.

**Question Identifier:** ExtendedQA-Q12
**Question:** I see a right-sided enlarged lymph node posterolateral to the anterior belly of the digastric muscles on a CT of the neck. What is the corresponding neck lymph node level?
**Reference Standard Answer:** Level Ib.
**Supspecialties:** Computed Tomography, Head and Neck.

**Question Identifier:** ExtendedQA-Q13
**Question:** I am looking at a CT angiography of the abdomen at the arterial phase. I see a complete defect of contrast in the infrarenal aorta extending into both common iliac arteries. What is this condition called?
**Reference Standard Answer:** Leriche syndrome.
**Supspecialties:** Computed Tomography, Emergency Radiology, Vascular Imaging.

**Question Identifier:** ExtendedQA-Q14
**Question:** I see a lesion in an MRI of the liver that is hyperintense on T2 relative to the surrounding liver parenchyma, but less than the intensity of water. Upon contrast administration, the lesion shows early peripheral nodular enhancement that is shifting centripetally on delayed images. What is the most likely diagnosis?
**Reference Standard Answer:** Hepatic hemangioma.
**Supspecialties:** Gastrointestinal, Magnetic Resonance Imaging.

**Question Identifier:** ExtendedQA-Q15
**Question:** What is the most commonly used chemotherapy agent in transcatheter arterial chemoembolisation of hepatocellular carcinomas?
**Reference Standard Answer:** Doxorubicin.
**Supspecialties:** Interventional Radiology.

**Question Identifier:** ExtendedQA-Q16
**Question:** I see an adrenal lesion in an MRI of the epigastric region. The lesion demonstrates a strong signal dropout on opposed-phase images. What is the most likely diagnosis?
**Reference Standard Answer:** Adrenal adenoma.
**Supspecialties:** Gastrointestinal, Magnetic Resonance Imaging.

**Question Identifier:** ExtendedQA-Q17
**Question:** I am looking at a contrast-enhanced CT of the abdomen on the arterial phase in a patient with an infrarenal abdominal aortic aneurysm that has been treated by EVAR. I see contrast agent in the aneurysmal sac, originating from a left-sided lumbar artery. What is the most likely diagnosis?
**Reference Standard Answer:** Endoleak type IIa.
**Supspecialties:** Computed Tomography, Vascular Imaging.

**Question Identifier:** ExtendedQA-Q18
**Question:** Strongly hyperdense lesions are noted in the segmental pulmonary arteries of the right lower lobe on a non-contrast CT of the thorax. The patient has a history of vertebroplasty. What kind of lesions could that be?
**Reference Standard Answer:** Pulmonary cement embolism.
**Supspecialties:** Chest, Computed Tomography, Musculoskeletal.

**Question Identifier:** ExtendedQA-Q19
**Question:** I see a lesion in an MRI of the liver that is slightly hyperintense on T2 and hypointense on T1 relative to the surrounding liver parenchyma. The lesion shows heterogenous arterial enhancement and a washout in the venous phase. The lesion shows a central scar on T2w imaging. What is the most likely diagnosis?
**Reference Standard Answer:** Fibrolamellar hepatocellular carcinoma.
**Supspecialties:** Gastrointestinal, Magnetic Resonance Imaging.

**Question Identifier:** ExtendedQA-Q20
**Question:** I see prominent endplate densities at multiple contiguous vertebral levels on a sagittal X-ray of the spine and a well-defined lesion of the sixth rib on a chest X-ray. What is the most likely underlying condition?
**Reference Standard Answer:** Hyperparathyroidism (Rugger-Jersey-Spine and 'brown tumor').
**Supspecialties:** Chest, Musculoskeletal.

**Question Identifier:** ExtendedQA-Q21
**Question:** After application of contrast agent into a percutaneous transhepatic biliary drainage, portal vein branches distal to the biliary entry of the drainage show contrast enhancement. What is the most likely diagnosis?
**Reference Standard Answer:** Portobiliary fistula.
**Supspecialties:** Gastrointestinal, Magnetic Resonance Imaging.

**Question Identifier:** ExtendedQA-Q22

**Question:** I see densely sclerotic endplates in a sagittal X-ray of the spine of a 14 year old patient. What is the most likely diagnosis?
**Reference Standard Answer:** Osteopetrosis.
**Supspecialties:** Musculoskeletal, Pediatric.

**Question Identifier:** ExtendedQA-Q23
**Question:** I see mixed opacified and unopacified blood in the IVC at and cranial of the renal veins after on a abdominal CT scan at an early contrast phase after i.v. contrast application. The infrarenal vena cava appears mostly unopacified. What is the most likely diagnosis?
**Reference Standard Answer:** Contrast flow phenomenon in inferior vena cava (artifact).
**Supspecialties:** Computed Tomography, Vascular Imaging.

**Question Identifier:** ExtendedQA-Q24
**Question:** I see a synovial mass on a knee MRI. The mass is T1 and T2 hyperintense. There is a concomitant joint effusion. What is the most likely diagnosis?
**Reference Standard Answer:** Lipoma arborescens.
**Supspecialties:** Magnetic Resonance Imaging, Musculoskeletal.

# Appendix S3

## Glossary of Technical Terms

**Tokens**:

In language models, *tokens* are small units of text (often words or subwords) used by the model to process, represent, or generate text. For example, the sentence "MRI is a diagnostic tool" might be split into tokens like [*"MRI", "is", "a", "diagnostic", "tool"*].

**Vector Representations**:

A *vector representation* (sometimes called an "embedding") is a way to translate text, images, or other data into a series of numerical values—often stored as a list or "vector"—such that similar content has similar numerical patterns. This allows algorithms to mathematically compare the semantic meaning of different inputs.

**Text Embedding**:

A *text embedding* is a type of vector representation specifically designed for linguistic content. It encodes each text chunk into a fixed-dimensional vector space such that texts with related meanings end up closer together in that space

**Vector Database**:

A *vector database* is a specialized data management system optimized for storing and retrieving vector representations. Traditional databases store and index discrete data types (like numbers or strings), whereas a vector database is designed to handle operations like *similarity search* (for instance, retrieving the closest vectors to a particular query vector).

# Appendix S4

**Time Comparison Between Conventional QA and RadioRAG-powered QA**

The time required for question answering (QA) varies depending on the model used and the retrieval approach. Below, we provide an estimate of the time differences between conventional QA and RadioRAG-powered QA, as observed in our experiments using the RSNA-RadioQA dataset.

**Conventional QA**:
For conventional QA, the time required is primarily dependent on the LLM's response time. For instance, when using GPT-4 through the OpenAI API, it took approximately 30 seconds per question to generate a response.

**RadioRAG-powered QA**:
RadioRAG introduces additional steps for retrieving and processing relevant context before querying the LLM, which increases the total time. This process consists of the following steps:

- **Key-phrase extraction**: Approximately 10 seconds, dependent on internet speed.

- **Searching Radiopaedia using key-phrases**: Approximately 30 seconds, dependent on both internet speed and the response time of the Radiopaedia website.

- **Database creation and embedding**: With the collection of around 25 articles, this step took approximately 40 seconds, influenced by the hardware resources available.

- **LLM response generation**: This step took approximately 35 seconds, similar to conventional QA.

In total, RadioRAG-powered QA took approximately 115 seconds per question using GPT-4, about four times longer than conventional QA (30 seconds).